\documentclass[letterpaper]{article} 
\usepackage{aaai2026}  
\usepackage{times}  
\usepackage{helvet}  
\usepackage{courier}  
\usepackage[hyphens]{url}  
\usepackage{graphicx} 
\usepackage{natbib}  
\usepackage{caption} 
\usepackage{algorithm}
\usepackage{algorithmic}
\usepackage{amsmath}
\usepackage{bm}
\usepackage{amsthm}
\usepackage{booktabs}       
\usepackage{amsfonts}       
\usepackage{nicefrac}       
\usepackage{microtype}      
\usepackage{xcolor}         
\usepackage{mathtools}
\usepackage{siunitx}  
\usepackage{multirow}

\usepackage{times}
\usepackage{helvet}
\usepackage{courier}
\usepackage{xcolor}

\usepackage{amsmath}
\usepackage{amssymb}
\usepackage{mathtools}
\usepackage{amsthm}

\usepackage[utf8]{inputenc} 
\usepackage{url}            
\usepackage{booktabs}       
\usepackage{amsfonts}       
\usepackage{nicefrac}       
\usepackage{microtype}      
\usepackage{xcolor}         

\usepackage{xcolor}         
\usepackage{algorithm}
\usepackage{algorithmic}
\usepackage{amsmath}
\usepackage{amssymb}
\usepackage{mathtools}
\usepackage{amsthm}
\usepackage{multicol}

\usepackage{caption}
\usepackage{graphicx}
\usepackage{enumitem}
\usepackage{subfigure}
\usepackage{bm}
\usepackage{array}
\usepackage{indentfirst}

\usepackage{lipsum}		
\usepackage{natbib}
\usepackage{appendix}
\usepackage{import}
\usepackage{bbding}
\usepackage{threeparttable}
\usepackage{dsfont}
\usepackage{mathrsfs}

\usepackage{graphicx}
\usepackage{multirow}
\usepackage{makecell}

\usepackage{minted} 
\usepackage{xcolor} 
\definecolor{bg}{rgb}{0.95,0.95,0.92} 

\newcommand{\eg}{{\em e.g.}}           
\newcommand{\ie}{{\em i.e.}}           

\theoremstyle{plain}
\newtheorem{theorem}{Theorem}[section]

\newtheorem{lemma}[theorem]{Lemma}

\theoremstyle{definition}

\theoremstyle{remark}

\usepackage{amsmath}
\usepackage{amssymb}
\usepackage{booktabs}

\RequirePackage{etex} 
\usepackage{cleveref}
\usepackage{listings}
\usepackage{mathtools}
\usepackage{autonum}
\usepackage{bibunits}

\DeclareMathOperator*{\argmin}{arg\,min}

\usepackage{newfloat}
\usepackage{listings}
\DeclareCaptionStyle{ruled}{labelfont=normalfont,labelsep=colon,strut=off} 
\floatstyle{ruled}
\newfloat{listing}{tb}{lst}{}
\floatname{listing}{Listing}
\title{Faster Game Solving via Asymmetry of Step Sizes}
\author{
    Linjian Meng,\textsuperscript{\rm 1}
    Tianpei Yang,\textsuperscript{\rm 1}\thanks{Corresponding Authors.}
    Youzhi Zhang,\textsuperscript{\rm 2{$*$}}
    Zhenxing Ge,\textsuperscript{\rm 1}
    Yang Gao\textsuperscript{\rm 1}
}
\affiliations{
    \textsuperscript{\rm 1} National Key Laboratory for Novel Software Technology, Nanjing University, Nanjing, China\\
    \textsuperscript{\rm 2} Centre for Artificial Intelligence and Robotics, Hong Kong Institute of Science \& Innovation, CAS\\

    menglinjian@smail.nju.edu.cn, tianpei.yang@nju.edu.cn, youzhi.zhang@cair-cas.org.hk,\\ zhenxingge@smail.nju.edu.cn, gaoy@nju.edu.cn 
    
}

\usepackage{bibentry}

\begin{document}

\maketitle

\begin{abstract}
Counterfactual Regret Minimization (CFR) algorithms are widely used to compute a Nash equilibrium (NE) in two-player zero-sum imperfect-information extensive-form games (IIGs). Among them, Predictive CFR$^+$ (PCFR$^+$) is particularly powerful, achieving an exceptionally fast empirical convergence rate via the prediction in many games. However, the empirical convergence rate of PCFR$^+$ would significantly degrade if the prediction is inaccurate, leading to unstable performance on certain IIGs. To enhance the robustness of PCFR$^+$, we propose Asymmetric PCFR$^+$ (APCFR$^+$), which employs an adaptive asymmetry of step sizes between the updates of implicit and explicit accumulated counterfactual regrets to mitigate the impact of the prediction inaccuracy on convergence. We present a theoretical analysis demonstrating why APCFR$^+$ can enhance the robustness. To the best of our knowledge, we are the first to propose the asymmetry of step sizes, a simple yet novel technique that effectively improves the robustness of PCFR$^+$. Then, to reduce the difficulty of implementing APCFR$^+$ caused by the adaptive asymmetry, we propose a simplified version of APCFR$^+$ called Simple APCFR$^+$ (SAPCFR$^+$), which uses a fixed asymmetry of step sizes to enable only a single-line modification compared to original PCFR$^+$. Experimental results on five standard IIG benchmarks and two heads-up no-limit Texas Hold’em (HUNL) Subagems show that (i) both APCFR$^+$ and SAPCFR$^+$ outperform PCFR$^+$ in most of the tested games, (ii) SAPCFR$^+$ achieves a comparable empirical convergence rate with APCFR$^+$, and (iii) our approach can be generalized to improve other CFR algorithms, \eg, Discount CFR (DCFR). 
\end{abstract}

\begin{links}
    \link{Code}{https://github.com/menglinjian/AAAI-2026-APCFRPlus}
\end{links}

\section{Introduction}

\begin{figure}[t]
    \centering 
    \subfigure{
    \includegraphics[width=1.0\linewidth]{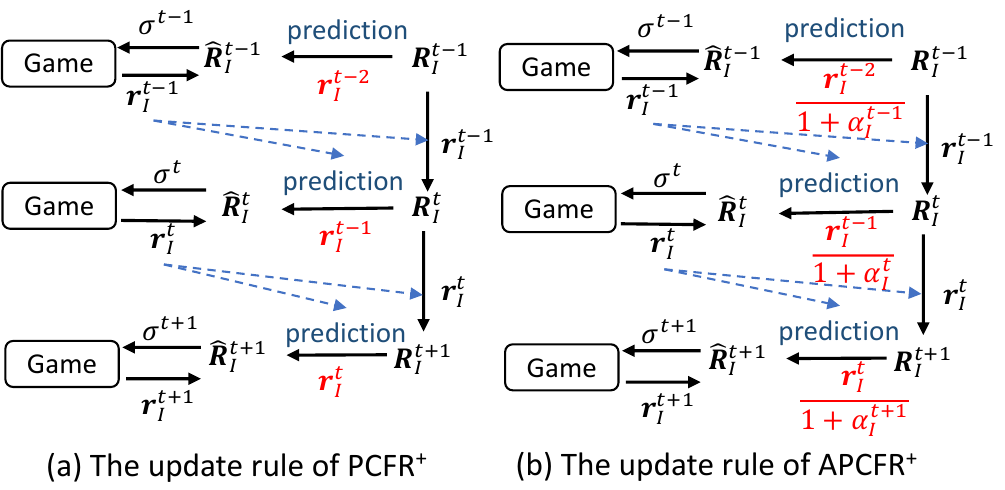}
    }  
    \caption{Comparison between PCFR$^+$ and APCFR$^+$, with differences highlighted in {red}. Note that the notation $t$ in $\alpha^t_I$ denotes iteration $t$, rather than an exponent.}
    \label{fig:update_rule_comparison}
\end{figure} 

\noindent Imperfect-information extensive-form games (IIGs) are foundational models to capture interactions among multiple agents in sequential settings with hidden information. IIGs are widely used to simulate real-world scenarios such as medical treatment~\citep{sandholm2015steering}, security games~\citep{lisy2016counterfactual}, cybersecurity~\citep{chen2017wireless}, and recreational games~\citep{brown2018superhuman, brown2019superhuman}. To address IIGs, a primary goal is to compute a Nash equilibrium (NE), where no player can unilaterally improve its payoff by deviating from the equilibrium.

As with much of the literature on solving IIGs, we focus on computing an NE in two-player zero-sum IIGs. The most widely used method for computing an NE in these IIGs is Counterfactual Regret Minimization (CFR) ~\citep{zinkevich2007regret,lanctot2009monte,tammelin2014solving,brown2019solving,farina2021faster,farina2019online,liu2021equivalence,liu2022power,meng2023efficient,farina2023regret,xu2022autocfr,xu2024minimizing,xu2024dynamic,zhang2024faster}, as evidenced by their success in superhuman game AIs~\citep{bowling2015heads,moravvcik2017deepstack,brown2018superhuman,brown2019superhuman,perolat2022mastering}. The key insight of CFR algorithms is to decompose the total regret over the game into a sum of counterfactual regrets associated within information sets (infosets) and employ a local regret minimizer to minimize counterfactual regrets within each infoset.

Many technologies have been proposed to improve the empirical convergence rate of CFR algorithms. For example, Counterfactual Regret Minimization$^+$ (CFR$^+$)~\citep{tammelin2014solving} replaces the local regret minimizer—Regret Matching (RM)~\citep{hart2000simple,gordon2006no}—used in vanilla CFR with Regret Matching$^+$ (RM$^+$). CFR$^+$ improves the empirical convergence rate by ensuring that the accumulated counterfactual regrets remain non-negative. Then, \citet{farina2021faster} introduce Predictive CFR$^+$ (PCFR$^+$), an improved variant of CFR$^+$. 



\begin{figure}[t]
    \centering 
    \subfigure{
    \includegraphics[width=0.9
    \linewidth]{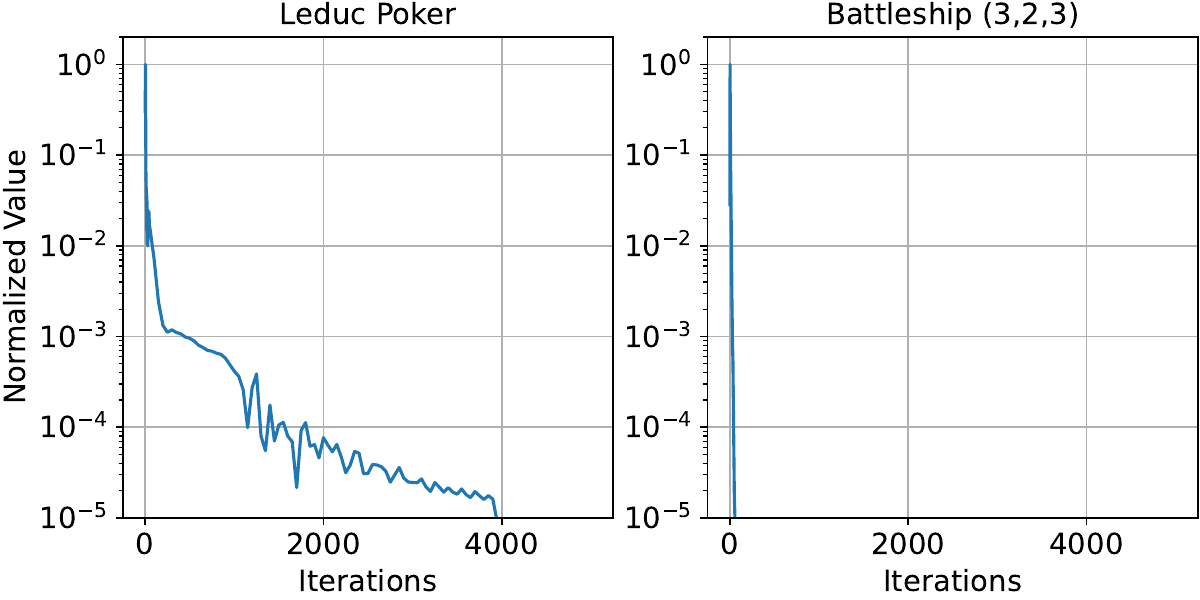}
    }
    \caption{Dynamics of inaccuracy in PCFR$^+$ between predicted and observed instantaneous counterfactual regrets in Leduc Poker and Battleship (3,2,3). This inaccuracy is related to the theoretical convergence rate of PCFR$^+$. The values on the y-axis are normalized to the range {[0, 1], which is displayed on a logarithmic scale.}}
\label{fig:dynamics_regret_bound_imm_gap_regret_gap_without_alpha_introduction}
\end{figure}

PCFR$^+$ significantly outperforms other CFR algorithms including CFR$^+$ in many IIGs by using the prediction. Specifically, PCFR$^+$ maintains two types of accumulated counterfactual regrets: the implicit and the explicit. As shown in \Cref{fig:update_rule_comparison}, at each iteration $t$, PCFR$^+$ uses the prediction and the observed instantaneous counterfactual regret $\bm{r}^t_I$ to derive the new explicit accumulated counterfactual regret $\hat{\bm{R}}^t_I$ and the new implicit counterfactual regret $\bm{R}^{t+1}_I$, respectively. If the prediction aligns with the observed instantaneous counterfactual regret $\bm{r}^t_I$, the theoretical convergence rate of PCFR$^+$ can be improved from $O(1/\sqrt{T})$ of CFR$^+$ to $O(1/T)$~\citep{farina2021faster}. However, PCFR$^+$ sets the instantaneous counterfactual regret $\bm{r}^{t-1}_I$ observed at iteration $t-1$ as the prediction at iteration $t$. {This operation may cause inaccurate prediction on certain IIGs, which harms the empirical convergence rate of PCFR$^+$.} As noted by \citet{farina2021faster}, PCFR$^+$ underperforms other CFR algorithms in Leduc Poker (Game [\textbf{O}] in \citet{farina2021faster}), yet significantly surpasses them in Battleship (3,2,3) (Game [\textbf{R}] in \citet{farina2021faster}). This aligns with the results in \Cref{fig:dynamics_regret_bound_imm_gap_regret_gap_without_alpha_introduction}: the gap between predicted and observed instantaneous counterfactual regret decreases slowly in Leduc Poker but diminishes rapidly in Battleship (3,2,3), validating our hypothesis.

To enhance the robustness of PCFR$^+$, we propose a novel variant of PCFR$^+$, termed Asymmetric PCFR$^+$ (APCFR$^+$). Similar to PCFR$^+$, APCFR$^+$ leverages the prediction to improve the convergence rate, but it mitigates the impact of the prediction inaccuracy on convergence. Specifically, as illustrated in \Cref{fig:update_rule_comparison}, APCFR$^+$ utilizes an adaptive asymmetry mechanism for step sizes between implicit and explicit accumulated counterfactual regret updates, which dynamically reduces the step size  when updating via the prediction. We prove that when the step size for updating the explicit accumulated counterfactual regret via the prediction at iteration $t$ is set to ${1}/{(1+\alpha^t)}$ for APCFR$^+$, where $\alpha^t \geq 0$ is a constant, the effect of the prediction inaccuracy on the convergence rate for APCFR$^+$ is reduced by a factor of $1 + \alpha^t$ compared to PCFR$^+$. Therefore, APCFR$^+$ mitigates the impact of the prediction inaccuracy on the convergence rate. Then, through the theoretical analysis of APCFR$^+$, we propose an automatic learning mechanism for $\alpha^t$, eliminating the need for fine-tuning parameters across different games. To the best of our knowledge, we are the first to propose the asymmetry of step sizes updating implicit and explicit accumulated counterfactual regrets. 

{To simplify the implementation of APCFR$^+$ caused by the automatic learning approach of $\alpha^t$,} we introduce a simplified version of APCFR$^+$, called Simple APCFR$^+$ (SAPCFR$^+$). Specifically, by analyzing the upper bounds of different terms within the theoretical guarantee of APCFR$^+$ (detailed at the beginning of Section \ref{subsec:SAPCFR+}), SAPCFR$^+$ sets $\alpha^t$ to $2$, ensuring SAPCFR$^+$ requires only a single-line modification to the original PCFR$^+$ code.

We conduct extensive experimental evaluations of APCFR$^+$ and SAPCFR$^+$ across five standard IIG benchmarks as well as two heads-up no-limit Texas Hold'em (HUNL) Subgames generated by the top poker agent, Libratus~\citep{brown2018superhuman}. The experiments demonstrate that APCFR$^+$ and SAPCFR$^+$ outperforms PCFR$^+$ in nearly all tested games and achieve an empirical convergence rate comparable to that of PCFR$^+$ in the remaining games. Moreover, we observe that SAPCFR$^+$ achieves comparable empirical convergence rate with APCFR$^+$. {Finally,  we can observe that our approach can be generalized to improve other CFR algorithms, \eg, Discount CFR (DCFR)~\citep{brown2019solving}.}

\section{Related Work}

\noindent We consider CFR algorithms~\citep{zinkevich2007regret,tammelin2014solving,brown2019solving,farina2021faster,farina2019online,liu2021equivalence,perolat2021poincare,liu2022power,meng2023efficient,farina2023regret,xu2022autocfr,xu2024minimizing,xu2024dynamic,zhang2024faster}, the most widely used method for learning an NE in two-player zero-sum IIGs, as evidenced by their success in superhuman game AIs~\citep{bowling2015heads,moravvcik2017deepstack,brown2018superhuman,brown2019superhuman,perolat2022mastering}.

The key insight of CFR algorithms is the decomposition of the regret over the game into the sum of counterfactual regrets associated with infosets. The vanilla CFR algorithm, introduced by \citet{zinkevich2007regret}, employs RM~\citep{hart2000simple} as the local regret minimizer. To improve the empirical convergence rate of CFR, it is common to design more effective local regret minimizers, as the selection of the local regret minimizers has a significant impact on the overall performance of the CFR algorithm. Examples include RM$^+$~\citep{bowling2015heads}, Discounted RM\ (DRM)~\citep{brown2019solving}, and PRM$^+$~\citep{farina2021faster}, which correspond to CFR$^+$, DCFR, and PCFR$^+$, respectively. PCFR$^+$ can demonstrate an extremely faster empirical convergence rate than other CFR variants. However, as shown in its original paper, PCFR$^+$ is outperformed by CFR$^+$ and DCFR even on standard IIG benchmarks like Leduc Poker.

To improve the robustness of PCFR$^+$, \citet{farina2023regret} propose Stable PCFR$^+$ and Smooth PCFR$^+$. These algorithms improve the robustness by addressing the instability, \ie, rapid strategy fluctuations across iterations, via ensuring the lower bound of the 1-norm of accumulated counterfactual regrets exceeds a positive constant. However, these algorithms never outperform PCFR$^+$ in terms of the empirical convergence rate even though they achieve a faster theoretical convergence rate than PCFR$^+$, as demonstrated in our experiments. APCFR$^+$ does not focus on addressing the instability, but instead aims to mitigate the impact of the prediction inaccuracy on the convergence to improve the robustness. In our experiments, APCFR$^+$ consistently outperforms Stable PCFR$^+$ and Smooth PCFR$^+$ in all tested games.

\section{Preliminaries}\label{sec:Preliminaries}
\noindent \textbf{Imperfect-information Extensive-form games (IIGs).} To model tree-form sequential decision-making problems with hidden information, a common used model is IIG~\citep{osborne2004introduction}. An IIG can be formulated as $G=\{\mathcal{N}, \mathcal{H}, P, A, \mathcal{I}, \{u_i\}\}$.  {Here,} $\mathcal{N} = \{0,1\}$ is the set of players. ``Nature” is also considered a player $c$ (representing chance) and chooses actions with a fixed known probability distribution. $\mathcal{H}$ is the set of all possible histories. For each history $h\in \mathcal{H}$, the function $P(h)$ represents the player {acting at} history $h${, and} $A(h)$ denotes the actions available at history $h$. To {account for} private information, the histories for each player $i$ are partitioned into a collection $\mathcal{I}_i$, {referred to as} information sets\ (infosets). For any {infoset} $I \in \mathcal{I}_i$, histories $h,h'\in I$ are indistinguishable to player $i$. The notation $\mathcal{I}$ denotes $\mathcal{I} = \{ \mathcal{I}_i | i \in \mathcal{N}\}$. Thus, we have $P(I)=P(h)$, $A(I)=A(h), \forall h \in I$. The set of leaf nodes is denoted by $\mathcal{Z}$. For each leaf node $z$, there is a pair $(u_0(z), u_1(z)) \in [-1,\ 1]$ which denotes the payoffs for the min player\ (player 0) and the max player\ (player 1) respectively. In two-player zero-sum IIGs, $u_0(z) = -u_1(z), \forall z \in \mathcal{Z}$.


\textbf{Behavioral strategy.} This strategy $\sigma_i$ is defined on each infoset. For any infoset $I \in \mathcal{I}_i$, the probability for an action $a \in A(I)$ is denoted by $\sigma_i(I,a)$. We use $\sigma_i(I) = [\sigma_i(I,a)|a \in A(I)] \in \Delta^{|A(I)|}$ to denote the strategy at infoset $I$, where $\Delta^{|A(I)|}$ is a $(|A(I)|-1)$-dimension simplex. If every player follows the strategy profile $\sigma=[ \sigma_0; \sigma_1]$ and reaches infoset $I$, the reaching probability is denoted by $\pi^{{\sigma}}(I)$. The probability contribution of player $i$ is $\pi_i^{\sigma}(I)$, while for players other than $i$, denoted as $-i$, the contribution is $\pi_{-i}^{\sigma}(I)$. In IIGs, $u_i(\sigma) = u_i(\sigma_i, \sigma_{-i})=\sum_{z \in \mathcal{Z}}u_i(z)\pi^{\sigma}(z)$.

\textbf{Nash equilibrium (NE).} NE denotes a rational behavior where no player can benefit by unilaterally deviating from the equilibrium. For any player, her strategy is the best-response to the strategies of others. Formally, for any NE strategy profile $\sigma^{*}$ and $i \in \mathcal{N}$, it holds that $u_i(\sigma^{*}_i, \sigma^{*}_{-i}) \geq u_i(\sigma_i, \sigma^{*}_{-i})$ for all $\sigma$. A widely used metric to measure the distance from the given strategy profile $\bm{x}$ to NE is the exploitability, which is defined as $\epsilon(\sigma) =\sum_{i \in \mathcal{N}} \max_{\sigma_i^{\prime}} (u_i(\sigma_i^{\prime}, \sigma_{-i}) - u_i(\sigma_i, \sigma_{-i}))/|\mathcal{N}|$. 

\textbf{Computing an NE via regret minimization algorithms.} 
To compute an NE in IIGs, a common used method is regret minimization algorithms~\citep{rakhlin2013online,rakhlin2013optimization,hazan2016introduction,joulani2017modular}. For any sequence of strategies $\sigma_i^1, \cdots, \sigma_i^T$ of player $i$, player $i$'s regret is $R^{T}_i =  \max_{\sigma_i} \sum^{T}_{t=1} u_i(\sigma_i, \sigma^t_{-i}) - \sum^{T}_{t=1}  u_i(\sigma^t_i, \sigma^t_{-i})$. Regret minimization algorithms are algorithms ensuring $R^{T}_i$ grows sublinearly. If each player follows a regret minimization algorithm, then their average strategy converges to the set of the NE in two-player zero-sum IIGs. Formally, assume the regret of each player $i$ is $R^{T}_i$, then it holds that 
\begin{equation}\label{eq:exploitability and regret bound}
    \epsilon(\bar{\sigma}) = \epsilon(\bar{\sigma}_0, \bar{\sigma}_1) \leq {\sum_{i \in \mathcal{N}} R^{T}_i}/{(|\mathcal{N}|T)},\ 
\end{equation}
where $\bar{\sigma}_i(I) = \sum_{t=1}^T \pi^{\sigma^t}_i(I){\sigma^t_i}(I)/\sum_{t=1}^T \pi^{\sigma^t}_i(I)$. 

\textbf{Counterfactual Regret Minimization (CFR) framework.} This framework~\citep{zinkevich2007regret, farina2019online, liu2021equivalence} is designed to compute an NE of two-player zero-sum IIGs. Instead of directly minimizing the global regret $R^{T}_i$, it decomposes the regret to each {infoset} and independently minimizes the local regret {within each infoset}. Let $\sigma^t$ be the strategy profile at iteration $t$. This framework computes the counterfactual value at infoset $I$ for action $a$ as $
    v^{\sigma^t}(I,a) = \sum_{h \in I} \sum_{z \in \mathcal{Z}_{ha}} \pi^{\sigma^t}_{-i} (h) \pi^{\sigma^t}(ha, z) u_i(z),\
$
where $\pi^{\sigma^t}(ha, z)$ denotes the probability from $ha$ to $z$ if all players play according to $\sigma^t$ and $\mathcal{Z}_{ha}$ is the set of the leaf nodes that are reachable after choosing action $a$ at history $h$. For any infoset $I$, the counterfactual regret is $
    R^{T}(I) =  \max_{a \in A(I)} \sum_{t=1}^{T} v^{\sigma^t}(I,a) - \sum_{t=1}^{T} \sum_{a' \in A(I)} \sigma^t_i (I,a') v^{\sigma^t}(I,a').
$
The regret over the game $R^{T}_i = \max_{\sigma_i} \sum^{T}_{t=1} u_i(\sigma_i, \sigma^t_{-i}) - \sum^{T}_{t=1}  u_i(\sigma^t_i, \sigma^t_{-i})$ is less than the sum of the counterfactual regrets within infosets: $
R^{T}_i \leq \sum_{I \in \mathcal{I}_i} R^{T}(I).
$
So any regret minimization algorithms can be used as the local regret minimizer to minimize the regret $R^{T}(I)$ over each infoset to minimize the regret $R^{T}_i$.

\textbf{Predictive Counterfactual Regret Minimization$^+$ (PCFR$^+$).} PCFR$^+$~\citep{farina2021faster} is a powerful CFR algorithm, which significantly outperforms other CFR algorithm in many IIGs. PCFR$^+$ employs Predictive RM$^+$ (PRM$^+$)~\citep{farina2021faster} as its local regret minimizer, with its key insight is to use the prediction. Specifically, as shown \Cref{fig:update_rule_comparison}, at each iteration $t$, PCFR$^+$ maintains implicit and explicit accumulated counterfactual regrets: ${\bm{R}}^{t}_I$ and $\hat{\bm{R}}^{t-1}_I$. Firstly, PCFR$^+$ makes a prediction and uses this prediction to derive new explicit accumulated counterfactual regrets $\hat{\bm{R}}^{t}_I$ from ${\bm{R}}^{t}_I$. Then, PCFR$^+$ observes the instantaneous counterfactual regret $\bm{r}^{t}_I$ by following the strategy $\sigma^{t}$ defined by  $\hat{\bm{R}}^{t}_I$. Lastly, $\bm{r}^{t}_I$ is subsequently used to derive ${\bm{R}}^{t+1}_I$ from ${\bm{R}}^{t}_I$. If the prediction aligns with the observed instantaneous counterfactual regret $\bm{r}^{t}_I$, \citet{farina2021faster} show that the theoretical convergence of PCFR$^+$ can be improved from $O(1/\sqrt{T})$ of CFR$^+$ to $O(1/T)$. As tested in \citet{farina2021faster}, using the instantaneous counterfactual regret $\bm{r}^{t-1}_I$ observed at the previous iteration $t-1$ as the prediction is both simple and effective. Therefore, in practice, PCFR$^+$ uses $\bm{r}^{t-1}_I$ as the prediction at iteration $t$. Formally, at each iteration $t$ and for each infoset $I \in \mathcal{I}$, PCFR$^+$ updates its strategy according to
\begin{equation}\label{eq:update-rule-PRM+}
\begin{aligned}
    & 
    \hat{\bm{R}}^{t}_I = [{\bm{R}}^{t}_I + \bm{r}^{t-1}_I]^+,\quad
    {\bm{R}}^{t+1}_I = [{\bm{R}}^{t}_I + \bm{r}^{t}_I]^+,\\
    & \sigma^{t}_i(I) = \frac{[\hat{\bm{R}}^{t}_I]^+}{\Vert [\hat{\bm{R}}^{t}_I]^+ \Vert_1}= \frac{\hat{\bm{R}}^{t}_I}{\Vert \hat{\bm{R}}^{t}_I \Vert_1},\\
\end{aligned}
\end{equation}
where $i=P(I)$, ${\bm{R}}^{1}_I = \bm{0}$, and the forth equality comes from the fact that $\hat{\bm{R}}^{t}_I \geq \bm{0}$.

\section{Methodology}\label{sec:our_methods}

\noindent PCFR$^+$ leverages the prediction to accelerate the empirical convergence rate. However, when the prediction is inaccurate, its empirical convergence rate may decrease significantly, leading to unstable performance on certain IIGs. To enhance the robustness of PCFR$^+$, we propose Asymmetric PCFR$^+$ (APCFR$^+$), which mitigates the impact of the prediction inaccuracy on the convergence rate via the adaptive asymmetry of step sizes. We then provide a theoretical analysis for APCFR$^+$ to demonstrate the reason why it enhances the robustness. To simplify the implementation of APCFR$^+$ due to the adaptive asymmetry, we propose Simple APCFR$^+$ (SAPCFR$^+$), using a constant asymmetry to guarantee that it can be implemented with a single-line modification compared to PCFR$^+$.

\subsection{Asymmetric PCFR$^+$ (APCFR$^+$)}\label{subsec:APCFR+}

\noindent {To mitigate the impact of the prediction inaccuracy on convergence of PCFR$^+$, APCFR$^+$ adaptively reduces the step size when updating via the prediction, i.e., when updating the explicitly accumulated counterfactual regret. In other words, APCFR$^+$ exploits the adaptive asymmetry of step sizes between the updates of the implicit and explicit ones.} Formally, at iteration $t$ and infoset $I$, the update rule of APCFR$^+$ is
\begin{equation}\label{eq:update-rule-P2PCFR+}
\begin{aligned}
    & 
    \hat{\bm{R}}^{t}_I = [{\bm{R}}^{t}_I + \frac{1}{1+\alpha^t_I}\bm{r}^{t-1}_I]^+,\quad
    {\bm{R}}^{t+1}_I = [{\bm{R}}^{t}_I + \bm{r}^{t}_I]^+,\\  
    &\sigma^{t}_i(I) = \frac{[\hat{\bm{R}}^{t}_I]^+}{\Vert [\hat{\bm{R}}^{t}_I]^+ \Vert_1}= \frac{\hat{\bm{R}}^{t}_I}{\Vert \hat{\bm{R}}^{t}_I \Vert_1},\\
\end{aligned}
\end{equation}
where $i=P(I)$, ${\bm{R}}^{1}_I = \bm{0}$, and $\bm{r}^{0}_I = \bm{0}$. {The comparison between the update rules of PCFR$^+$ and APCFR$^+$ has been shown in \Cref{fig:update_rule_comparison}.} {In the rest of this subsection, we first present the regret upper bound for APCFR$^+$ with respect to any $\alpha^t_I$, as stated in Theorem \ref{thm:regret bound of P2PCFR}. According to the discussion about Theorem \ref{thm:regret bound of P2PCFR}, we show why APCFR$^+$ can enhance the robustness of PCFR$^+$ by mitigating the impact of the prediction inaccuracy on the convergence rate. Lastly, we discuss how to automatically learn $\alpha^t_I$ from the regret bound shown in Theorem \ref{thm:regret bound of P2PCFR}.}


\begin{theorem}\label{thm:regret bound of P2PCFR}
[Proof is in Appendix \ref{sec:proof thm:regret bound of P2PCFR}].
Assuming that $T$ iterations of APCFR$^+$ with any $\alpha^t_I \geq 0$ are conducted, the counterfactual regret at any infoset $I \in \mathcal{I}$ is bound by
\[
R^{T}(I) \leq  \sqrt{{\sum^T_{t=1}\left(\frac{\Vert {\bm{r}}^{t}_I -\bm{r}^{t-1}_I \Vert^2_2}{1+\alpha^t_I} + \alpha^t_I \Vert {\bm{R}}^{t+1}_I -{\bm{R}}^t_I \Vert^2_2\right)}}.
\]
\end{theorem}

\textbf{Why the asymmetry mechanism is effective.} To assess the effectiveness of the asymmetry mechanism for step sizes in decreasing the regret upper bound (improving the convergence rate), we show the upper bound of $\Vert {\bm{r}}^{t}_I - \bm{r}^{t-1}_I \Vert^2_2$ is four times than that of $\Vert {\bm{R}}^{t+1}_I - {\bm{R}}^t_I \Vert^2_2$. Firstly, we introduce Lemma \ref{thm:bound of two accumulated regret}.

\begin{lemma}\label{thm:bound of two accumulated regret}
[Adapted from Lemma 11 of \citet{wei2020linear}].
Assume that $T$ iterations of APCFR$^+$ with any $\alpha^t_I \geq 0$ are conducted. Then for any infoset $I \in \mathcal{I}$ and $t \geq 1$, we have
\[
\Vert {\bm{R}}^{t+1}_I - {\bm{R}}^t_I \Vert^2_2 \leq \Vert {\bm{r}}^{t}_I \Vert^2_2.
\]
\end{lemma}

Assume that for any infoset \( I \in \mathcal{I} \) and \( t \geq 1 \), \( \Vert {\bm{r}}^{t}_I \Vert^2_2 \leq E \). Then, from Lemma \ref{thm:bound of two accumulated regret}, we have
\begin{equation}\label{eq:Sec-4-2-0}
\begin{aligned}
    \Vert {\bm{R}}^{t+1}_I - {\bm{R}}^t_I \Vert^2_2 \leq E.
\end{aligned}
\end{equation}
Similarly, for \( \Vert {\bm{r}}^{t}_I - \bm{r}^{t-1}_I \Vert^2_2 \), we have
\begin{equation}\label{eq:Sec-4-2-1}
\begin{aligned}
    \Vert {\bm{r}}^{t}_I - \bm{r}^{t-1}_I \Vert^2_2 \leq 4E.
\end{aligned}
\end{equation}
{In experiments, we also analyze the values of two terms
{$
\sum^T_{t=1} \Vert {\bm{r}}^{t}_I -\bm{r}^{t-1}_I \Vert^2_2 \
\text{and} 
\sum^T_{t=1} \Vert {\bm{R}}^{t+1}_I -{\bm{R}}^t_I \Vert^2_2,
$} 
for both PCFR$^+$ and our algorithms (Figures \ref{fig:dynamics_regret_bound_imm_gap_regret_gap_without_alpha} and \ref{fig:dynamics_regret_bound_subgames_imm_gap_regret_gap_without_alpha}). Among all algorithms, we observe that the value of {$\sum^T_{t=1} \Vert {\bm{r}}^{t}_I -\bm{r}^{t-1}_I \Vert^2_2$} is at least three times than that of {$\sum^T_{t=1} \Vert {\bm{R}}^{t+1}_I - {\bm{R}}^t_I \Vert^2_2$}. This indicates that introducing the term {$\sum^T_{t=1} \alpha^t_I \Vert {\bm{R}}^{t+1}_I - {\bm{R}}^t_I \Vert^2_2$} and modifying the term {$\sum^T_{t=1} \Vert {\bm{r}}^{t}_I - \bm{r}^{t-1}_I \Vert^2_2$} to {$\sum^T_{t=1} \frac{\Vert {\bm{r}}^{t}_I - \bm{r}^{t-1}_I \Vert^2_2}{1+\alpha^t_I}$}, can reduce the regret upper bound. Furthermore, compared to PCFR$^+$, both of these two terms are smaller in our algorithms, further decreasing the regret upper bound. Then, we evaluate the values of 
{$
{\sum^T_{t=1} ( \frac{\Vert {\bm{r}}^t_I - {\bm{r}}^{t-1}_I \Vert^2_2}{1 + \alpha^t_I} + \alpha^t_I \Vert {\bm{R}}^{t+1}_I - {\bm{R}}^t_I \Vert^2_2 )},
$} for both PCFR$^+$ and our algorithms (Figures \ref{fig:dynamics_regret_bound} and \ref{fig:dynamics_regret_bound_subgames}). In all games, the value of {$\sum^T_{t=1} ( \frac{\Vert {\bm{r}}^t_I - {\bm{r}}^{t-1}_I \Vert^2_2}{1 + \alpha^t_I}+ \alpha^t_I \Vert {\bm{R}}^{t+1}_I - {\bm{R}}^t_I \Vert^2_2 )$} is consistently smaller in our algorithms than in PCFR$^+$. See more details in Appendix \ref{sec:Additional Experiments}.
}



\textbf{An alternative regret upper bound of APCFR$^+$.} {Notably, Theorem \ref{thm:regret bound of P2PCFR} does not conflict the upper regret bound of CFR$^+$ (where $\alpha^t_I \to \infty$), as it provides a larger upper regret bound than the original CFR$^+$ upper bound. By altering the proof method, we get {$R^{T}(I) \leq  \sqrt{\sum^T_{t=1} \Vert {\bm{r}}^{t}_I - \frac{1}{1+\alpha^t_I} \bm{r}^{t-1}_I \Vert^2_2}$}, as shown in Theorem \ref{thm:regret bound of P2PCFR-2} (detailed in Appendix \ref{sec:Another Regret Bound of APCFR}). By setting $\alpha^t_I \to \infty$, the original bound of CFR$^+$ can be recovered. Additionally, for PCFR$^+$ (where $\alpha^t_I \to 0$), the bound in Theorem \ref{thm:regret bound of P2PCFR} is identical to the one in its original version (the result in Theorem 3 of the original PCFR$^+$ version can be easily improved to the bound presented in Theorem \ref{thm:regret bound of P2PCFR}). The reason why we employ Theorem \ref{thm:regret bound of P2PCFR} in the main text rather than Theorem \ref{thm:regret bound of P2PCFR-2} is that the regret bound in Theorem \ref{thm:regret bound of P2PCFR-2} is typically larger than that in Theorem \ref{thm:regret bound of P2PCFR}, as demonstrated in Appendix \ref{sec:Additional Experiments} (Figures \ref{fig:dynamics_regret_bound}, \ref{fig:dynamics_regret_bound_subgames}, \ref{fig:dynamics_regret_bound2}, and \ref{fig:dynamics_regret_bound2_subgames}).}

\textbf{Automatic learning approach for $\alpha^t_I$.} To eliminate the fine-tuning of $\alpha^t_I$, we propose an automatic learning approach for $\alpha^t_I$. From Theorem \ref{thm:regret bound of P2PCFR}, we have
\begin{equation}\label{eq:Sec-4-0}
%
\begin{aligned}
     R^{T}(I) & \leq   \sqrt{{\sum^T_{t=1}\left( \frac{\Vert \bm{r}^{t}_I - \bm{r}^{t-1}_I \Vert^2_2}{1+\alpha^t_I} + \alpha^t_I \Vert \bm{R}^{t+1}_I - \bm{R}^t_I \Vert^2_2 \right)}}  \\
     & \leq  \sqrt{{\sum^T_{t=1}\left( \frac{\Vert \bm{r}^{t}_I - \bm{r}^{t-1}_I \Vert^2_2}{\alpha^t_I} + \alpha^t_I \Vert \bm{R}^{t+1}_I - \bm{R}^t_I \Vert^2_2 \right)}} .
\end{aligned}
\end{equation}
To minimize the right-hand side of the last inequality, we can set
$%
    \alpha^t_I = \sqrt{\frac{\Vert \bm{r}^{t}_I - \bm{r}^{t-1}_I \Vert^2_2}{\Vert \bm{R}^{t+1}_I - \bm{R}^t_I \Vert^2_2}}.
$
However, this is not feasible, as we need $\alpha^t_I$ to compute $\bm{r}^{t}_I$. Therefore, we adopt an alternative approach:
\begin{equation}\label{eq:Sec-4-2}
\begin{aligned}
   \alpha^t_I = \min\left( \sqrt{\frac{\sum_{\tau=1}^{t-1}\Vert \bm{r}^{\tau}_I - \bm{r}^{\tau-1}_I \Vert^2_2}{\sum_{\tau=1}^{t-1} \Vert \bm{R}^{\tau+1}_I - \bm{R}^{\tau}_I \Vert^2_2}}, \alpha_{max} \right).
\end{aligned}
\end{equation}
Note that the parameter in $\alpha_{max}$ in Eq. (\ref{eq:Sec-4-2}) is included solely to ensure that the bound in Theorem \ref{thm:regret bound of P2PCFR} remains finite. In this paper, we directly set it as $5$ to reduce the cost of hyperparameter tuning. In practice, we rarely observed $\alpha^t_I$ reaching 5 (Figures \ref{fig:dynamics_alpha} and \ref{fig:dynamics_alpha_subgames}).

\subsection{Simple APCFR$^+$ (SAPCFR$^+$)}\label{subsec:SAPCFR+}

\noindent {To simplify the implementation of APCFR$^+$ caused by the automatic learning approach of $\alpha^t$,} we introduce SAPCFR$^+$, which is implemented with a single-line modification to the PCFR$^+$ code. {Specifically, SAPCFR$^+$ sets $\alpha^t_I=2$.} The key insight of setting $\alpha^t_I=2$ lies in the fact that the upper bound of $\Vert {\bm{R}}^{t+1}_I - {\bm{R}}^t_I \Vert^2_2$ is only a quarter of the upper bound of $\Vert {\bm{r}}^{t}_I - \bm{r}^{t-1}_I \Vert^2_2$, as shown in Eq. (\ref{eq:Sec-4-2-0}), and Eq. (\ref{eq:Sec-4-2-1}).

Specifically, combining Theorem \ref{thm:regret bound of P2PCFR}, Eq. (\ref{eq:Sec-4-2-0}), and Eq. (\ref{eq:Sec-4-2-1}), in the worst case, we obtain
\begin{equation}\label{eq:Sec-4-2-2}
%
\begin{aligned}
     R^{T}(I) & \leq   \sqrt{{\sum^T_{t=1}\left( \frac{\Vert \bm{r}^{t}_I - \bm{r}^{t-1}_I \Vert^2_2}{1+\alpha^t_I} + \alpha^t_I \Vert \bm{R}^{t+1}_I - \bm{R}^t_I \Vert^2_2 \right)}}   \\
     &\leq  \sqrt{{\sum^T_{t=1}\left( \frac{4E}{1+\alpha^t_I} + \alpha^t_I E \right)}} .
\end{aligned}
\end{equation}
It is evident that when $\alpha^t_I = 0$, i.e., for PCFR$^+$, the worst-case counterfactual regret upper bound is
\begin{equation}\label{eq:Sec-4-2-3}
\begin{aligned}
     R^{T}(I) & \leq   \sqrt{\sum^T_{t=1}  4E} .
\end{aligned}
\end{equation}
From the facts that (i)
$2$ minimizes {${{\sum^T_{t=1}( {4E}/{\alpha^t_I} + \alpha^t_I E )}}$} for any positive $E$ and (ii) {${{\sum^T_{t=1}( {4E}/{(1+\alpha^t_I)} + \alpha^t_I E )}} \leq {{\sum^T_{t=1}( {4E}/{\alpha^t_I} + \alpha^t_I E )}}$}, we can set $\alpha^t_I = 2$, which implies the counterfactual regret is bound by
\begin{equation}\label{eq:Sec-4-2-3}
\begin{aligned}
     R^{T}(I) & \leq   \sqrt{\sum^T_{t=1} \left( \frac{4E}{1+2} + 2E \right)} =  \sqrt{\sum^T_{t=1}  \frac{10E}{3}} .
\end{aligned}
\end{equation}
Clearly, setting $\alpha^t_I = 2$ results in a lower regret upper bound than PCFR$^+$. Therefore, for SAPCFR$^+$, we set $\alpha^t_I = 2$ for all $t \geq 1$. Formally, at each iteration $t$, SAPCFR$^+$ updates its strategy at each infoset $I \in \mathcal{I}$ according to the following update rule:
\begin{equation}\label{eq:update-rule-S2PCFR+}
\begin{aligned}
    & \hat{\bm{R}}^{t}_I = [ {\bm{R}}^{t}_I + \frac{1}{3} \bm{r}^{t-1}_I ]^+,\quad {\bm{R}}^{t+1}_I = [ {\bm{R}}^{t}_I + \bm{r}^{t}_I ]^+,\\
    &\sigma^{t}_i(I) = \frac{[\hat{\bm{R}}^{t}_I]^+}{\Vert [\hat{\bm{R}}^{t}_I]^+ \Vert_1}= \frac{\hat{\bm{R}}^{t}_I}{\Vert \hat{\bm{R}}^{t}_I \Vert_1},\\
\end{aligned}
\end{equation}
where $i=P(I)$, ${\bm{R}}^{1}_I = \bm{0}$, and $\bm{r}^{0}_I = \bm{0}$.

\section{Experiments}\label{sec:experiments}

\begin{figure*}[t!]
    \centering 
    \subfigure{
    \includegraphics[width=1\linewidth]{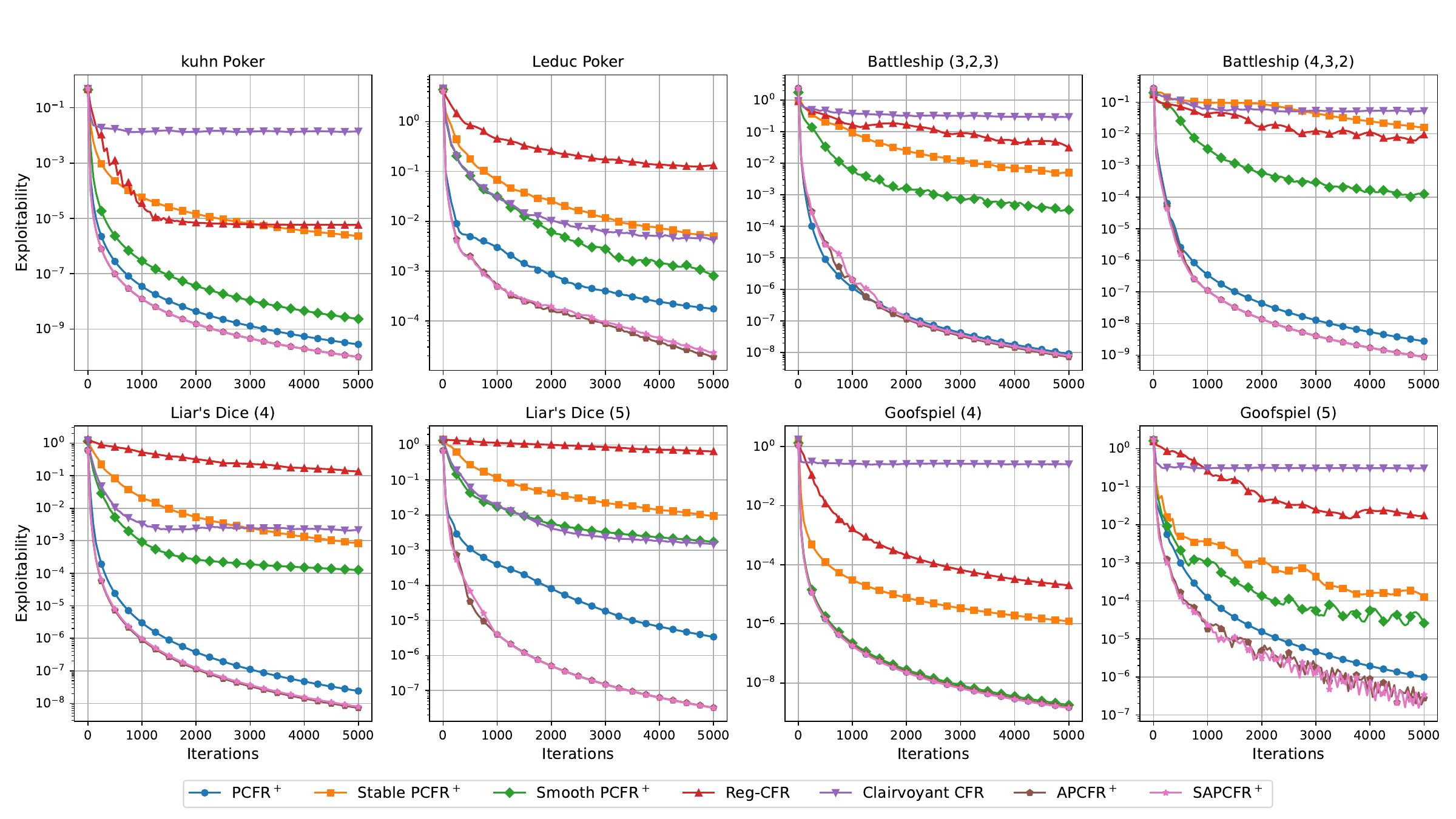}
    }
    \caption{Empirical convergence rates of the tested algorithms in standard commonly used IIG benchmarks. In all plots, the x-axis is the number of iterations, and the y-axis is exploitability, displayed on a logarithmic scale. Liar’s Dice ($x$) represents that every player is given a die with $x$ sides. Goofspiel ($x$) denotes that each player is dealt $x$ cards. Battleship ($x,y,z$) implies the size of the grid is $x\times y$, and the number of shots is $z$.
}
\label{fig:PSPCFR+-with-other-algorithms}
\end{figure*}


\noindent \textbf{Configurations.} 
We now evaluate the empirical convergence rates of APCFR$^+$ and SAPCFR$^+$ by comparing them to PCFR$^+$, Stable PCFR$^+$, Smooth PCFR$^+$, Reg-CFR~\citep{liu2022power}, and Clairvoyant CFR~\citep{farina2023regret}. Stable PCFR$^+$ and Smooth PCFR$^+$ are advanced PCFR$^+$ variants. Reg-CFR and Clairvoyant CFR achieve theoretical convergence rates of $O(1/T^{\frac{3}{4}})$ and $O(1/T)$, respectively, while that of other algorithms is $O(1/\sqrt{T})$. Following the settings in PCFR$^+$, we employ alternating updates for both APCFR$^+$ and SAPCFR$^+$. For Stable PCFR$^+$, Smooth PCFR$^+$, and Reg-CFR, we apply alternating updates, as described in their original paper or open-source code. Clairvoyant CFR does not utilize alternating updates, in accordance with its original design. For all algorithms, we utilize quadratic averaging. For all compared algorithms, we adopt the hyperparameters as suggested in their respective original versions. Details on the size of the tested games are in Appendix \ref{sec:Additional Experiments} (\Cref{tab:Number of Information Sets of Games}). The experiments are conducted on a machine equipped with a Xeon(R) Gold 6444Y CPU and 256 GB of memory.

\textbf{Empirical convergence rates in standard IIG benchmarks.} We now present the empirical convergence rates across five standard IIG benchmarks, \eg, Kuhn Poker, Leduc Poker, Goofspiel Poker, Liar's Dice, and Battleship. These games are implemented using OpenSpiel~\citep{lanctot2019openspiel}. The algorithm implementations are based on LiteEFG~\citep{liu2024liteefgefficientpythonlibrary}, as LiteEFG provides approximately 100 times speedup compared to the default implementation in OpenSpiel. The results are in Figure \ref{fig:PSPCFR+-with-other-algorithms}. For most of the tested games, except for Battleship (3,2,3) and Goofspiel (4), APCFR$^+$ and SAPCFR$^+$, significantly outperform all baselines. Even in Battleship (3,2,3) and Goofspiel (4), APCFR$^+$ and SAPCFR$^+$ outperform all algorithms except PCFR$^+$. Remarkably, they exhibit performance comparable to PCFR$^+$, reaching similar levels of exploitability after 5000 iterations. Based on the experimental results in Appendix \ref{sec:Additional Experiments} (Figures \ref{fig:dynamics_regret_bound_imm_gap_regret_gap_without_alpha} and \ref{fig:dynamics_regret_bound_subgames_imm_gap_regret_gap_without_alpha}), we observe that {in the games where our algorithms perform similar to PCFR$^+$, such as Battleship (3,2,3) and Goofspiel (4), PCFR$^+$ also exhibits a rapid decrease in the inaccuracy between the predicted and observed instantaneous counterfactual regrets (detailed discussions are in Appendix \ref{sec:Additional Experiments}).} Furthermore, the performance gap between APCFR$^+$ and SAPCFR$^+$ is relatively small. Specifically, APCFR$^+$ only outperforms SAPCFR$^+$ in Leduc Poker, Battleship (3,2,3), and Liar's Dice (5). This small performance gap means that in practical applications, SAPCFR$^+$ can be directly used due to its ease of implementation and a faster empirical convergence rate compared to PCFR$^+$. Regarding Stable PCFR$^+$ and Smooth PCFR$^+$, we find that they significantly underperform PCFR$^+$. Reg-CFR and Clairvoyant CFR significantly underperform relative to other algorithms.


\textbf{Empirical convergence rates in HUNL Subgames.} To assess the performance of APCFR$^+$ and SAPCFR$^+$ in addressing real-world games, we also conduct evaluations in HUNL Subgames, which are considerably larger than standard IIG benchmarks. Despite the presence of code related to HUNL Subgames in Openspiel, we have not successfully executed it. Therefore, we utilize HUNL Subgames implemented by Poker RL~\citep{steinberger2019pokerrl}. More precisely, our code is based on the code from~\citet{xu2024minimizing}. The code in~\citet{xu2024minimizing} supports only Subgame 3 and Subgame 4, so we conduct experiments solely on these two HUNL Subgames. We do not compare Reg-CFR and Clairvoyant CFR in HUNL Subgames, as they perform significantly worse than other CFR algorithms, even in standard IIG benchmarks. The results are shown in Table \ref{tab:PSPCFR+-with-other-algorithms-HUNL Subgames}: APCFR+ and SAPCFR+ consistently outperform all baselines in both subgames.

\textbf{Running times.} To validate the efficiency of APCFR$^+$ and SAPCFR$^+$, we compare their running time with that of PCFR$^+$ under the same number of iterations (\ie, 5000). The experimental results are in Appendix \ref{sec:Additional Experiments} (Table \ref{tab:runing time}). The running time of APCFR$^+$ is slightly higher compared to PCFR$^+$, primarily due to the additional $\alpha^t_I$ learning process in APCFR$^+$. However, the running time of SAPCFR$^+$ is nearly identical to that of PCFR$^+$, as the only difference between their implementations is a single line of code, which does not alter the computational complexity. Notably, the computational complexity remains exactly the same, even with no change in the constant factors. 

\begin{table*}[t]  
\centering %
\begin{tabular}{l@{\hskip 4pt}ccccc}  
\toprule  
& PCFR$^+$ & Stable PCFR$^+$  & Smooth PCFR$^+$ & APCFR$^+$ & SAPCFR$^+$ \\
\midrule  
Subgame 3 & \texttt{1.44e-3} & \texttt{1.41e-3} (\textcolor{red}{-2.1\%}) & \texttt{1.42e-3} (\textcolor{red}{-1.4\%}) & \texttt{1.02e-3} (\textcolor{red}{-29.2\%}) & \texttt{9.44e-4} (\textcolor{red}{-34.4\%})   \\
Subgame 4 & \texttt{1.04e-3} & \texttt{9.77e-4} (\textcolor{red}{-5.3\%}) & \texttt{1.02e-3} (\textcolor{red}{-1.9\%}) & \texttt{7.53e-4} (\textcolor{red}{-27.6\%}) & \texttt{7.83e-4} (\textcolor{red}{-24.7\%})  \\
\bottomrule  
\end{tabular}  
\caption{Final exploitability for the tested algorithms in HUNL Subgames. Values in \textcolor{red}{red} indicate percentages relative to PCFR$^+$.}  
\label{tab:PSPCFR+-with-other-algorithms-HUNL Subgames}  
\end{table*}  

\begin{table*}[!ht]   
    \centering  %
    \begin{tabular}{cccc}  
    \toprule  
     & Leduc Poker (5) & Leduc Poker (9) & Leduc Poker (13) \\
    \midrule  
    DCFR & \texttt{2.79e-5} & \texttt{1.27e-5} & \texttt{1.09e-5} \\
    PCFR$^+$ & \texttt{2.69e-5} & \texttt{5.21e-5} & \texttt{3.15e-5} \\
    APCFR$^+$ & \texttt{4.80e-6} (\textcolor{red}{-82.1\%}) & \texttt{4.03e-5} (\textcolor{red}{-22.6\%}) & \texttt{1.45e-5} (\textcolor{red}{-54.0\%}) \\
    SAPCFR$^+$ & \texttt{3.49e-6} (\textcolor{red}{-87.0\%}) & \texttt{4.07e-5} (\textcolor{red}{-21.9\%}) & \texttt{1.42e-5} (\textcolor{red}{-55.0\%}) \\
    DCFR$^+$ & \texttt{1.15e-5} (\textcolor{blue}{-58.8\%}) & \texttt{6.41e-6} (\textcolor{blue}{-49.5\%}) & \texttt{8.56e-6} (\textcolor{blue}{-21.6\%}) \\
    APDCFR$^+$ & \texttt{3.69e-6} (\textcolor{red}{-86.3\%}, \textcolor{blue}{-86.7\%}) & \texttt{3.42e-6} (\textcolor{red}{-93.4\%}, \textcolor{blue}{-73.1\%}) & \texttt{3.02e-6} (\textcolor{red}{-90.4\%}, \textcolor{blue}{-72.3\%}) \\
    \bottomrule  
    \end{tabular}   
    \caption{The final exploitability for DCFR, PCFR$^+$, APCFR$^+$, SAPCFR$^+$, DCFR$^+$, and APDCFR$^+$ in Leduc Poker variants. Values in \textcolor{red}{red} indicate percentages of PCFR$^+$ variants relative to PCFR$^+$, and values in \textcolor{blue}{blue} indicate percentages of DCFR variants relative to DCFR. Notably, APDCFR$^+$ can serve as a variant of both PCFR$^+$ and DCFR.}  
    \label{tab:leduc poker variants exploitability-classical-cfr}  
\end{table*}

\textbf{Dynamics of $\alpha^t_I$ in APCFR$^+$.} To study the behavior of $\alpha^t_I$, we analyze its dynamics, as shown in Appendix \ref{sec:Additional Experiments} (Figures \ref{fig:dynamics_alpha} and \ref{fig:dynamics_alpha_subgames}). We observe that $\alpha^t_I$ experiences a rapid increase during the initial phase but ceases to grow after approximately 100 iterations. This might be due to that the values of $\Vert {\bm{r}}^t_I - {\bm{r}}^{t-1}_I \Vert^2_2$ and $\Vert {\bm{R}}^{t+1}_I - {\bm{R}}^t_I \Vert^2_2$ are significantly larger in the initial phase than at later stages (as also observed in Figures \ref{fig:dynamics_regret_bound_imm_gap_regret_gap_without_alpha} and \ref{fig:dynamics_regret_bound_subgames_imm_gap_regret_gap_without_alpha}). More details are in Appendix \ref{sec:Additional Experiments}.

{\textbf{Dynamics of 
$
\sum_{t=1}^{T} \Vert \bm{r}^t_I - \bm{r}^{t-1}_I \Vert^2_2$, $\sum_{t=1}^{T} \Vert \bm{R}^{t+1}_I - \bm{R}^t_I \Vert^2_2$, $\sum_{t=1}^{T} ( \frac{\Vert \bm{r}^t_I - \bm{r}^{t-1}_I \Vert^2_2}{1 + \alpha^t_I} + \alpha^t_I \Vert \bm{R}^{t+1}_I - \bm{R}^t_I \Vert^2_2 )$, and $\sum^T_{t=1} \Vert {\bm{r}}^{t}_I - \frac{{\bm{r}}^{t-1}_I }{{1 + \alpha^t_I}}\Vert^2_2$.} To evaluate the regret bound presented in our theoretical analysis, we examine the dynamics of these terms, as demonstrated in Appendix \ref{sec:Additional Experiments}. Specifically, the dynamics of $\sum_{t=1}^{T} \Vert \bm{r}^t_I - \bm{r}^{t-1}_I \Vert^2_2$ and $\sum_{t=1}^{T} \Vert \bm{R}^{t+1}_I - \bm{R}^t_I \Vert^2_2$ are presented in Figures \ref{fig:dynamics_regret_bound_imm_gap_regret_gap_without_alpha} and \ref{fig:dynamics_regret_bound_subgames_imm_gap_regret_gap_without_alpha}. Similarly, Figures \ref{fig:dynamics_regret_bound} and \ref{fig:dynamics_regret_bound_subgames} present the dynamics of $\sum_{t=1}^{T} \left( \frac{\Vert \bm{r}^t_I - \bm{r}^{t-1}_I \Vert^2_2}{1 + \alpha^t_I} + \alpha^t_I \Vert \bm{R}^{t+1}_I - \bm{R}^t_I \Vert^2_2 \right)$. Additionally, Figures \ref{fig:dynamics_regret_bound2} and \ref{fig:dynamics_regret_bound2_subgames} depict the dynamics of $\sum^T_{t=1} \Vert \bm{r}^t_I - \frac{\bm{r}^{t-1}_I }{1 + \alpha^t_I}\Vert^2_2$. This experimental results show that  $\Vert {\bm{r}}^t_I - {\bm{r}}^{t-1}_I \Vert^2_2$ are larger than $\Vert {\bm{R}}^{t+1}_I - {\bm{R}}^t_I \Vert^2_2$, which confirms that APCFR$^+$ effectively reduces the impact of $\Vert {\bm{R}}^{t+1}_I - {\bm{R}}^t_I \Vert^2_2$ by increasing the weights on $\Vert {\bm{R}}^{t+1}_I - {\bm{R}}^t_I \Vert^2_2$. For the first three terms, their values are smaller in APCFR$^+$ and SAPCFR$^+$ compared to PCFR$^+$, implying a lower regret bound in Theorem \ref{thm:regret bound of P2PCFR}. However, the value of $\sum^T_{t=1} \Vert {\bm{r}}^{t}_I - \frac{{\bm{r}}^{t-1}_I }{{1 + \alpha^t_I}}\Vert^2_2$ significantly exceeds that of $\sum_{t=1}^{T} ( \frac{\Vert \bm{r}^t_I - \bm{r}^{t-1}_I \Vert^2_2}{1 + \alpha^t_I} + \alpha^t_I \Vert \bm{R}^{t+1}_I - \bm{R}^t_I \Vert^2_2 )$, indicating that the regret bound in Theorem \ref{thm:regret bound of P2PCFR-2} is extremely higher than that in Theorem \ref{thm:regret bound of P2PCFR}. Thus, we use Theorem \ref{thm:regret bound of P2PCFR} in the main text instead of Theorem \ref{thm:regret bound of P2PCFR-2}.}

\textbf{Empirical convergence rates of APCFR$^+$ with an alternative learning approach for $\alpha^t$.} We also experiment with a different learning approach for $\alpha^t$, other than the one in Eq. (\ref{eq:Sec-4-2}), \eg, $\alpha^t_I = \min\left( \sqrt{\frac{\max_{\tau \in [t-1]}\Vert \bm{r}^{\tau}_I - \bm{r}^{\tau-1}_I \Vert^2_2}{\max_{\tau \in [t-1]} \Vert \bm{R}^{\tau+1}_I - \bm{R}^{\tau}_I \Vert^2_2}}, \alpha_{max} \right)$, where we also set $\alpha_{max}=5$ as did in Eq. (\ref{eq:Sec-4-2}) to reduce the cost of hyperparameter tuning. The results in Appendix \ref{sec:Additional Experiments} (Figure \ref{fig:PSPCFR+ V2-with-other-algorithms} and \Cref{fig:PSPCFR+ V2-with-other-algorithms-HUNL Subgames}) indicate that this approach performs similarly to the one in Eq. (\ref{eq:Sec-4-2}).

\textbf{Comparison with other classical CFR algorithms and the generalization of our approach.} In addition to the CFR algorithms that have already been compared, we also compare APCFR$^+$ and SAPCFR$^+$ with the classic CFR algorithms: CFR, CFR$^+$, and DCFR. Initially, we conducted experiments using standard IIG benchmarks and HUNL Subgames (\Cref{fig:PSPCFR+-with-other-algorithms-classical-cfr} and \Cref{tab:PSPCFR+-with-other-algorithms-HUNL Subgames-classical-cfr}), where APCFR$^+$ and SAPCFR$^+$ consistently outperformed CFR and CFR$^+$ across all games. However, in poker games like Leduc Poker and HUNL Subgames, APCFR$^+$ and SAPCFR$^+$ did not surpass DCFR. Notably, our algorithms and DCFR are not mutually exclusive and can be combined effectively. The core innovation of our algorithms—the asymmetry of step sizes—can be integrated with DCFR, which involves discounting prior iterations when calculating accumulated regrets. Therefore, we propose APDCFR$^+$ by combining APCFR$^+$ with DCFR (details of APDCFR$^+$ are in Appendix \ref{sec:details of APDCFR}). In addition to CFR, CFR$^+$, DCFR, APCFR$^+$, and SAPCFR$^+$, we also compare APDCFR$^+$ with DCFR$^+$~\citep{xu2024minimizing}, which is an advanced variant of DCFR. Experimental results, detailed in Appendix \ref{sec:Additional Experiments} (\Cref{tab:PSPCFR+-with-other-algorithms-HUNL Subgames-classical-cfr}), demonstrate that APDCFR$^+$ achieves a substantially faster empirical convergence rate compared to the other evaluated algorithms.

{Additionally, to further evaluate the performance of DCFR, PCFR$^+$, APCFR$^+$, SAPCFR$^+$, DCFR$^+$, and APDCFR$^+$ in poker games, we conduct tests on various Leduc Poker variants that are used in the original PCFR$^+$ paper~\citep{farina2021faster}. Specifically, we test on Leduc Poker with ranks of 5, 9, or 13. We denote these Leduc Poker variants as Leduc Poker ($x$), where $x$ represents the number of ranks, noting that the original Leduc Poker has 3 ranks. The results, in Table \ref{tab:leduc poker variants exploitability-classical-cfr}, demonstrate that APCFR$^+$ and SAPCFR$^+$ consistently outperform PCFR$^+$ across all Leduc Poker variants. Notably, the degree to which APCFR$^+$ and SAPCFR$^+$ surpass PCFR$^+$ does not depend on the size of the game. Specifically, the smallest improvement of APCFR$^+$ and SAPCFR$^+$ over PCFR$^+$ occurs in Leduc Poker (9), where the reduction in exploitability is less than half of the reduction observed in Leduc Poker (13). Moreover, the results indicate that DCFR does not consistently outperform PCFR$^+$. For instance, in Leduc Poker (5), the performance of DCFR is inferior to that of PCFR$^+$. More importantly, APDCFR$^+$ consistently outperforms all other algorithms across each Leduc Poker variant tested, except in Leduc Poker (5), where it slightly underperforms compared to SAPCFR$^+$.}

\section{Conclusions}\label{sec:conclusion}

\noindent We propose a novel variant of PCFR$^+$ called APCFR$^+$, which employs the adaptive asymmetry of step sizes in the updates of implicit and explicit accumulated counterfactual regrets to improve the robustness of PCFR$^+$. We also introduce SAPCFR$^+$, requiring only a single line modification to PCFR$^+$. Experimental results validate that APCFR$^+$ and SAPCFR$^+$ exhibit a faster empirical convergence rate than PCFR$^+$. To our knowledge, we are the first to propose the asymmetry of step sizes in the updates of implicit and explicit accumulated counterfactual regrets, a simple yet novel technique that effectively improves the robustness of PCFR$^+$. Moreover, the techniques used in other CFR$^+$ algorithms are compatible with our algorithm, {which shows the generalization of our approach. For example, for DCFR, by using our approach, we propose APDCFR$^+$, which significantly outperforms DCFR} in poker games. Future work involves designing more effective $\alpha^t$ learning approaches to further enhance the empirical convergence rate.

\section*{Acknowledgements}
\noindent This work is supported in part by the National Natural Science Foundation of China under Grants 62192783 and 62506157, the Jiangsu Science and Technology Major Project BG2024031, the Fundamental Research Funds for the Central Universities (14380128), the Collaborative Innovation Center of Novel Software Technology and Industrialization, and the InnoHK funding.

\bibliography{references}

\appendix

\onecolumn

\section{Proof of Theorem \ref{thm:regret bound of P2PCFR}}\label{sec:proof thm:regret bound of P2PCFR}

\begin{proof}
    To prove Theorem \ref{thm:regret bound of P2PCFR}, we use the equivalence between RM$^+$ and Online Mirror Descent (OMD) proposed by \citet{farina2021faster}. We first introduce OMD. OMD is a traditional regret minimization algorithm~\citep{nemirovskij1983problem}. Let $\bm{\ell}^t \in \mathbb{R}^d$, $\bm{x}^t \in \mathcal{D}$, and let $\psi : \mathcal{D} \to \mathbb{R}^{d}_{\geq 0} = \{\bm{y}| \bm{y}\in \mathbb{R}^{d}, \bm{y} \geq \bm{0}\}$ be a 1-strongly convex differentiable regularizer with respect to some norm $\|\cdot\|$, OMD generates the decisions via 
    \begin{equation}\label{eq:OMD}
    %
      \bm{x}^{t+1} := \arg\min_{\bm{x}^{\prime} \in \mathcal{D}} \left\{ \langle \bm{\ell}^t, \bm{x}^{\prime} \rangle + \frac{1}{\eta} \mathcal{B}_{\psi}(\bm{x}^{\prime} \parallel \bm{x}^t) \right\},
    \end{equation}
    where $\mathcal{B}_{\psi} (\bm{u}, \bm{v}) = \psi(\bm{u}) - \psi(\bm{v}) - \langle \nabla \psi(\bm{v}), \bm{u} - \bm{v}\rangle$ is the Bregman divergence associated with $\psi(\cdot)$.

    From the analysis in Section D of \citet{farina2021faster}, by setting $\psi(\cdot)$ as the quadratic regularizer $\frac{1}{2}\Vert \cdot \Vert^2_2$, the update rule APCFR$^+$ at infoset $I$ can be written as 
    \begin{equation}\label{eq:update-rule-PRM+-GF}
    %
    \begin{aligned}
        & \hat{\bm{\theta}}^{t}_I \in \argmin_{\bm{\theta}_I \in \mathbb{R}^{|A(I)|}_{\geq 0}} \{ \langle  -\frac{1}{1+\alpha^t_I}\bm{r}^{t-1}_I, \bm{\theta}_I\rangle + \frac{1}{\eta} \mathcal{B}_{\psi} (\bm{\theta}_I, \bm{\theta}^{t}_I) \},\\
        & {\bm{\theta}}^{t+1}_I  \in \argmin_{\bm{\theta}_I \in \mathbb{R}^{|A(I)|}_{\geq 0}} \{ \langle  -\bm{r}^{t}_I, \bm{\theta}_I\rangle + \frac{1}{\eta} \mathcal{B}_{\psi} (\bm{\theta}_I, \bm{\theta}^{t}_I) \},
    \end{aligned}
    \end{equation}
    where $\eta > 0$ is a constant and $\sigma^t_i(I) = \frac{\hat{\bm{\theta}}^{t}_I}{\Vert \hat{\bm{\theta}}^{t}_I \Vert_1}$. Note that if $\bm{\theta}^{0}_I = \bm{0}$ for all $I \in \mathcal{I}$, then for any $\eta$, the strategy profile sequence $\{\sigma^1,\sigma^2,\cdots, \sigma^{T}\}$ generated by APCFR$^+$ is same and $\eta \bm{R}^t_I = \bm{\theta}^t_I$~\citep{farina2021faster}. 
    

    \begin{lemma}\label{lemma 4 of farina2021faster}
    [Lemma 4 of \citet{farina2021faster}]
    Let $\mathcal{D} \subseteq \mathbb{R}^d$ be closed and convex, let $\bm{\ell}^t \in \mathbb{R}^d$, $\bm{x}^t \in \mathcal{D}$, and let $\psi : \mathcal{D} \to \mathbb{R}_{\geq 0}$ be a 1-strongly convex differentiable regularizer with respect to some norm $\|\cdot\|$. Then,
    \[ 
    \bm{x}^{t+1} \in \arg\min_{\bm{x} \in \mathcal{D}} \left\{ \langle \bm{\ell}^t, \bm{x} \rangle + \frac{1}{\eta} \mathcal{B}_{\psi}(\bm{x} \parallel \bm{x}^t) \right\}
    \]
    is well defined (that is, the minimizer exists and is unique), and for all $\bm{x}^{\prime} \in \mathcal{D}$ satisfies the inequality
    \[ 
    %
    \thinmuskip=0mu
\medmuskip=0mu
\thickmuskip=0mu
\spaceskip=-0pt
    \langle \bm{\ell}^t, \bm{x}^{t+1} - \bm{x}^{\prime} \rangle \leq \frac{1}{\eta} \left( \mathcal{B}_{\psi}(\bm{x}^{\prime} , \bm{x}^t) - \mathcal{B}_{\psi}(\bm{x}^{\prime} , \bm{x}^{t+1}) - \mathcal{B}_{\psi}(\bm{x}^{t+1} , \bm{x}^t) \right).
    \]
    \end{lemma}

    Considering the second line of Eq. (\ref{eq:update-rule-PRM+-GF}), and using Lemma \ref{lemma 4 of farina2021faster} with $\bm{x}^t={\bm{\theta}}^t_I$, $\bm{x}^{t+1}={\bm{\theta}}^{t+1}_I$, $\bm{x}^{\prime} = \sigma_i(I)$ and $\bm{\ell}^t = -{\bm{r}}^{t}_I$, we have
    \begin{equation}\label{eq:sec:prf:lem:update-rule-SExRM+-inequality-0}
    %
    \thinmuskip=0.5mu
    \medmuskip=0.5mu
    \thickmuskip=0.5mu
    \spaceskip=0.5pt
    \begin{aligned}
         & \langle  -{\bm{r}}^{t}_I, {\bm{\theta}}^{t+1}_I - \sigma_i(I) \rangle \leq \frac{1}{\eta} \left(\mathcal{B}_{\psi}(\sigma_i(I), {\bm{\theta}}^t_I) - \mathcal{B}_{\psi}(\sigma_i(I), {\bm{\theta}}^{t+1}_I) - \mathcal{B}_{\psi}({\bm{\theta}}^{t+1}_I, {\bm{\theta}}^t_I) \right) \\
         \Leftrightarrow 
         & \langle  -{\bm{r}}^{t}_I, {\bm{\theta}}^{t+1}_I - \hat{\bm{\theta}}^t_I + \hat{\bm{\theta}}^t_I- \sigma_i(I) \rangle  \leq \frac{1}{\eta} \left( \mathcal{B}_{\psi}(\sigma_i(I), {\bm{\theta}}^t_I) - \mathcal{B}_{\psi}(\sigma_i(I), {\bm{\theta}}^{t+1}_I) - \mathcal{B}_{\psi}({\bm{\theta}}^{t+1}_I, {\bm{\theta}}^t_I) \right).
    \end{aligned}
    \end{equation}
    Similarly, considering the first line of Eq. (\ref{eq:update-rule-PRM+-GF}), and using Lemma \ref{lemma 4 of farina2021faster} with $\bm{x}^t={\bm{\theta}}^t_I$, $\bm{x}^{t+1}=\hat{\bm{\theta}}^t_I$, $\bm{x}^{\prime} = {\bm{\theta}}^{t+1}_I$ and $\bm{\ell}^t = -\bm{r}^{t-1}_I$, we get
    \begin{equation}\label{eq:sec4-temp-9}
    %
    \begin{aligned}
    & \frac{1}{1+\alpha^t_I}\langle  -\bm{r}^{t-1}_I, \hat{\bm{\theta}}^t_I - {\bm{\theta}}^{t+1}_I \rangle  \leq  \frac{1}{\eta} \left( \mathcal{B}_{\psi}({\bm{\theta}}^{t+1}_I, {\bm{\theta}}^t_I) - \mathcal{B}_{\psi}({\bm{\theta}}^{t+1}_I, \hat{\bm{\theta}}^t_I)  - \mathcal{B}_{\psi}(\hat{\bm{\theta}}^t_I, {\bm{\theta}}^t_I)  \right) \\
    \Leftrightarrow 
        & \langle  -\bm{r}^{t-1}_I, \hat{\bm{\theta}}^t_I - {\bm{\theta}}^{t+1}_I \rangle  \leq  \frac{1+\alpha^t_I}{\eta} \left( \mathcal{B}_{\psi}({\bm{\theta}}^{t+1}_I, {\bm{\theta}}^t_I) - \mathcal{B}_{\psi}({\bm{\theta}}^{t+1}_I, \hat{\bm{\theta}}^t_I)  - \mathcal{B}_{\psi}(\hat{\bm{\theta}}^t_I, {\bm{\theta}}^t_I)  \right).
    \end{aligned}
    \end{equation}
    Summing up Eq. (\ref{eq:sec:prf:lem:update-rule-SExRM+-inequality-0}) with Eq. (\ref{eq:sec4-temp-9}), we have
    \begin{equation}\label{eq:sec4-temp-10}
    %
    \begin{aligned}
        & \langle  -{\bm{r}}^{t}_I, {\bm{\theta}}^{t+1}_I - \hat{\bm{\theta}}^t_I + \hat{\bm{\theta}}^t_I- \sigma_i(I) \rangle + \langle  -\bm{r}^{t-1}_I, \hat{\bm{\theta}}^t_I - {\bm{\theta}}^{t+1}_I \rangle \\
        \leq & \frac{1}{\eta} \left(
        \mathcal{B}_{\psi}(\sigma_i(I), {\bm{\theta}}^t_I) - \mathcal{B}_{\psi}(\sigma_i(I), {\bm{\theta}}^{t+1}_I) - \mathcal{B}_{\psi}({\bm{\theta}}^{t+1}_I, {\bm{\theta}}^t_I) \right)  +  
        \\ & \quad \quad \quad
        \frac{1+\alpha^t_I}{\eta} \left(\mathcal{B}_{\psi}({\bm{\theta}}^{t+1}_I, {\bm{\theta}}^t_I)- \mathcal{B}_{\psi}({\bm{\theta}}^{t+1}_I, \hat{\bm{\theta}}^t_I) - \mathcal{B}_{\psi}({\hat{\bm{\theta}}}^{t}_I, {\bm{\theta}}^t_I)  \right),
    \end{aligned}
    \end{equation}
    which implies
    \begin{equation}\label{eq:sec4-temp-10-1}
    \thinmuskip=0mu
    \medmuskip=0mu
    \thickmuskip=0mu
    \spaceskip=-0pt
    \begin{aligned}
        & \langle  -{\bm{r}}^{t}_I,  \hat{\bm{\theta}}^t_I- \sigma_i(I) \rangle \\
        \leq & \langle  {\bm{r}}^{t}_I, {\bm{\theta}}^{t+1}_I - \hat{\bm{\theta}}^t_I \rangle + \langle  \bm{r}^{t-1}_I, \hat{\bm{\theta}}^t_I - {\bm{\theta}}^{t+1}_I \rangle + 
        \frac{1}{\eta} \left(         \mathcal{B}_{\psi}(\sigma_i(I), {\bm{\theta}}^t_I) - \mathcal{B}_{\psi}(\sigma_i(I), {\bm{\theta}}^{t+1}_I) - \mathcal{B}_{\psi}({\bm{\theta}}^{t+1}_I, {\bm{\theta}}^t_I) \right)  + 
        \\ & \quad \quad \quad
        \frac{1+\alpha^t_I}{\eta} \left(\mathcal{B}_{\psi}({\bm{\theta}}^{t+1}_I, {\bm{\theta}}^t_I)- \mathcal{B}_{\psi}({\bm{\theta}}^{t+1}_I, \hat{\bm{\theta}}^t_I)  - \mathcal{B}_{\psi}(\hat{\bm{\theta}}^{t}_I, {\bm{\theta}}^t_I)  \right)\\
        \leq & \langle  {\bm{r}}^{t}_I - \bm{r}^{t-1}_I , {\bm{\theta}}^{t+1}_I - \hat{\bm{\theta}}^t_I \rangle + 
        \frac{1}{\eta} \left(
        \mathcal{B}_{\psi}(\sigma_i(I), {\bm{\theta}}^t_I) - \mathcal{B}_{\psi}(\sigma_i(I), {\bm{\theta}}^{t+1}_I-  \mathcal{B}_{\psi}({\bm{\theta}}^{t+1}_I, {\bm{\theta}}^t_I)  \right)  + 
        \\ & \quad \quad \quad
        \frac{1+\alpha^t_I}{\eta} \left(\mathcal{B}_{\psi}({\bm{\theta}}^{t+1}_I, {\bm{\theta}}^t_I)- \mathcal{B}_{\psi}({\bm{\theta}}^{t+1}_I, \hat{\bm{\theta}}^t_I)  - \mathcal{B}_{\psi}(\hat{\bm{\theta}}^{t}_I, {\bm{\theta}}^t_I)  \right).
    \end{aligned}
    \end{equation}
    From the facts that (i) ${\bm{r}}^{t}_I = \bm{v}^{\sigma^{t}}(I) - \langle \bm{v}^{\sigma^{t}}(I), \sigma^{t}_i(I)\rangle \bm{1} $ and (ii) $\sigma^{t}_i(I) = \frac{[\hat{\bm{\theta}}^{t}_I]^+}{\Vert [\hat{\bm{\theta}}^{t}_I]^+ \Vert_1}= \frac{\hat{\bm{\theta}}^{t}_I}{\Vert \hat{\bm{\theta}}^{t}_I \Vert_1}$, we have
    \begin{equation}\label{eq:sec4-temp-12}
    %
    \begin{aligned}
         \langle  -{\bm{r}}^{t}_I, \hat{\bm{\theta}}^t_I- \sigma_i(I) \rangle 
        = -& \langle \langle \bm{v}^{\sigma^{t}}(I), \sigma^{t}_i(I)\rangle \bm{1} - \bm{v}^{\sigma^{t}}(I), \sigma_i(I) - \hat{\bm{\theta}}^t_I \rangle \\
        = -& \langle \langle \bm{v}^{\sigma^{t}}(I), \frac{\hat{\bm{\theta}}^{t}_I}{\Vert \hat{\bm{\theta}}^{t}_I \Vert_1} \rangle \bm{1} - \bm{v}^{\sigma^{t}}(I), \sigma_i(I) - \hat{\bm{\theta}}^t_I \rangle \\
        = -& \langle \langle \bm{v}^{\sigma^{t}}(I), \frac{\hat{\bm{\theta}}^{t}_I}{\Vert \hat{\bm{\theta}}^{t}_I \Vert_1} \rangle \bm{1} - \bm{v}^{\sigma^{t}}(I), \sigma_i(I) \rangle - \langle \langle \bm{v}^{\sigma^{t}}(I), \frac{\hat{\bm{\theta}}^{t}_I}{\Vert \hat{\bm{\theta}}^{t}_I \Vert_1}\rangle \bm{1} - \bm{v}^{\sigma^{t}}(I),  \hat{\bm{\theta}}^t_I \rangle  \\
         = -& \langle \langle \bm{v}^{\sigma^{t}}(I), \frac{\hat{\bm{\theta}}^{t}_I}{\Vert \hat{\bm{\theta}}^{t}_I \Vert_1} \rangle \bm{1} - \bm{v}^{\sigma^{t}}(I), \sigma_i(I) \rangle  - \langle \langle \bm{v}^{\sigma^{t}}(I), \frac{\hat{\bm{\theta}}^{t}_I}{\Vert \hat{\bm{\theta}}^{t}_I \Vert_1}\rangle \bm{1},  \hat{\bm{\theta}}^t_I \rangle  + \langle \bm{v}^{\sigma^{t}}(I),  \hat{\bm{\theta}}^t_I \rangle  \\
         = -& \langle \langle \bm{v}^{\sigma^{t}}(I), \frac{\hat{\bm{\theta}}^{t}_I}{\Vert \hat{\bm{\theta}}^{t}_I \Vert_1} \rangle \bm{1} - \bm{v}^{\sigma^{t}}(I), \sigma_i(I) \rangle  - \langle \bm{v}^{\sigma^{t}}(I), \frac{\hat{\bm{\theta}}^{t}_I}{\Vert \hat{\bm{\theta}}^{t}_I \Vert_1}\rangle \Vert \hat{\bm{\theta}}^{t}_I \Vert_1  + \langle \bm{v}^{\sigma^{t}}(I),  \hat{\bm{\theta}}^t_I \rangle  \\
        = -& \langle \langle \bm{v}^{\sigma^{t}}(I), \frac{\hat{\bm{\theta}}^{t}_I}{\Vert \hat{\bm{\theta}}^{t}_I \Vert_1} \rangle \bm{1} - \bm{v}^{\sigma^{t}}(I), \sigma_i(I) \rangle  - \langle \bm{v}^{\sigma^{t}}(I),  \hat{\bm{\theta}}^t_I \rangle  + \langle \bm{v}^{\sigma^{t}}(I),  \hat{\bm{\theta}}^t_I \rangle  \\
        = -& \langle \langle \bm{v}^{\sigma^{t}}(I), \frac{\hat{\bm{\theta}}^{t}_I}{\Vert \hat{\bm{\theta}}^{t}_I \Vert_1} \rangle \bm{1} - \bm{v}^{\sigma^{t}}(I), \sigma_i(I) \rangle \\
        = -&  \langle \bm{v}^{\sigma^{t}}(I), \sigma^{t}_i(I)- \sigma_i(I) \rangle = \langle \bm{v}^{\sigma^{t}}(I), \sigma_i(I) - \sigma^{t}_i(I)\rangle,
    \end{aligned}
    \end{equation}
    where the fifth line comes from $\langle\bm{1},\hat{\bm{\theta}}^t_I \rangle = \Vert \hat{\bm{\theta}}^t_I\Vert_1$, as well as the last line is from $\frac{\hat{\bm{\theta}}^{t}_I}{\Vert \hat{\bm{\theta}}^{t}_I \Vert_1} = \sigma^{t}_i(I)$ and $\langle \bm{1}, \sigma_i(I) \rangle = 1$ (as $\sigma_i(I) \in \Delta^{|A(I)|}$).
    In addition, we have
    \begin{equation}\label{eq:sec4-temp-13}
    %
    \begin{aligned}
         R^{T}(I) = & \max_{a \in A(I)} \sum_{t=1}^{T} v^{\sigma^t}(Ia) -  \sum_{t=1}^{T} \sum_{a \in A(I)} \sigma^t_i (Ia) v^{\sigma^t}(Ia) \\
        \leq & \max_{\sigma_i(I)} \sum_{t=1}^{T} \langle \bm{v}^{\sigma^{t}}(I), \sigma_i(I) - \sigma^{t}_i(I) \rangle.
    \end{aligned}
    \end{equation}
    Therefore, from Eq. (\ref{eq:sec4-temp-10-1}) and $\langle  -{\bm{r}}^{t}_I, \hat{\bm{\theta}}^t_I- \sigma_i(I) \rangle = \langle \bm{v}^{\sigma^{t}}(I), \sigma_i(I) - \sigma^{t}_i(I)\rangle$, we can bound 
    \begin{equation}\label{eq:sec4-temp-14}
        \thinmuskip=0mu
\medmuskip=0mu
\thickmuskip=0mu
\spaceskip=-0pt
    \begin{aligned}
    & \langle  {\bm{r}}^{t}_I - \bm{r}^{t-1}_I , {\bm{\theta}}^{t+1}_I - \hat{\bm{\theta}}^t_I \rangle + 
        \frac{1}{\eta} \left(         \mathcal{B}_{\psi}(\sigma_i(I), {\bm{\theta}}^t_I) - \mathcal{B}_{\psi}(\sigma_i(I), {\bm{\theta}}^{t+1}_I) - \mathcal{B}_{\psi}({\bm{\theta}}^{t+1}_I, {\bm{\theta}}^t_I) \right)  + 
        \\ & \quad \quad \quad
        \frac{1+\alpha^t_I}{\eta} \left(\mathcal{B}_{\psi}({\bm{\theta}}^{t+1}_I, {\bm{\theta}}^t_I)- \mathcal{B}_{\psi}({\bm{\theta}}^{t+1}_I, \hat{\bm{\theta}}^t_I)  - \mathcal{B}_{\psi}(\hat{\bm{\theta}}^{t}_I, {\bm{\theta}}^t_I)  \right),
    \end{aligned}
    \end{equation}
    to bound $R^{T}(I)$.

    

    For Eq. (\ref{eq:sec4-temp-14}), summing up from $t=1$ to $t=T$, we get
    \begin{equation}\label{eq:sec4-temp-15}
    \thinmuskip=0mu
\medmuskip=0mu
\thickmuskip=0mu
\spaceskip=-0pt
    \begin{aligned}
    & \sum^T_{t=1} \langle  {\bm{r}}^{t}_I - \bm{r}^{t-1}_I , {\bm{\theta}}^{t+1}_I - \hat{\bm{\theta}}^t_I \rangle +
        \sum^T_{t=1} \frac{1}{\eta} \left(
        \mathcal{B}_{\psi}(\sigma_i(I), {\bm{\theta}}^t_I) - \mathcal{B}_{\psi}(\sigma_i(I), {\bm{\theta}}^{t+1}_I) - \mathcal{B}_{\psi}({\bm{\theta}}^{t+1}_I, {\bm{\theta}}^t_I) \right)  + 
        \\ & \quad \quad \quad  
        \sum^T_{t=1} \frac{1+\alpha^t_I}{\eta} \left(\mathcal{B}_{\psi}({\bm{\theta}}^{t+1}_I, {\bm{\theta}}^t_I)- \mathcal{B}_{\psi}({\bm{\theta}}^{t+1}_I, \hat{\bm{\theta}}^t_I)  - \mathcal{B}_{\psi}(\hat{\bm{\theta}}^{t}_I, {\bm{\theta}}^t_I)  \right) \\
    \leq & \frac{1}{\eta}   
        \mathcal{B}_{\psi}(\sigma_i(I), {\bm{\theta}}^1_I) +  \sum^T_{t=1} \Vert {\bm{r}}^{t}_I -\bm{r}^{t-1}_I \Vert_2  \Vert {\bm{\theta}}^{t+1}_I - \hat{\bm{\theta}}^t_I \Vert_2  + 
        \\ & \quad \quad \quad
        \sum^T_{t=1} \frac{\alpha^t_I}{\eta}\mathcal{B}_{\psi}({\bm{\theta}}^{t+1}_I, {\bm{\theta}}^t_I)+\sum^T_{t=1} \frac{1+\alpha^t_I}{\eta} \left(- \mathcal{B}_{\psi}({\bm{\theta}}^{t+1}_I, \hat{\bm{\theta}}^t_I)  - \mathcal{B}_{\psi}(\hat{\bm{\theta}}^{t}_I, {\bm{\theta}}^t_I)  \right) \\
    \leq & \frac{1}{\eta}   
        \mathcal{B}_{\psi}(\sigma_i(I), {\bm{\theta}}^1_I) +  \sum^T_{t=1} \eta \frac{\Vert {\bm{r}}^{t}_I -\bm{r}^{t-1}_I \Vert^2_2}{2(1+\alpha^t_I)} + \sum^T_{t=1} (1+\alpha^t_I) \frac{\Vert {\bm{\theta}}^{t+1}_I - \hat{\bm{\theta}}^t_I \Vert_2}{2\eta}  + 
        \\ & \quad \quad \quad
        \sum^T_{t=1} \frac{\alpha^t_I}{\eta}\mathcal{B}_{\psi}({\bm{\theta}}^{t+1}_I, {\bm{\theta}}^t_I)+\sum^T_{t=1} \frac{1+\alpha^t_I}{\eta} \left(- \mathcal{B}_{\psi}({\bm{\theta}}^{t+1}_I, \hat{\bm{\theta}}^t_I)  - \mathcal{B}_{\psi}(\hat{\bm{\theta}}^{t}_I, {\bm{\theta}}^t_I)  \right) \\
    \leq & \frac{1}{\eta}   
        \mathcal{B}_{\psi}(\sigma_i(I), {\bm{\theta}}^1_I) +  \sum^T_{t=1} \eta \frac{\Vert {\bm{r}}^{t}_I -\bm{r}^{t-1}_I \Vert^2_2}{2(1+\alpha^t_I)}  + \sum^T_{t=1} \frac{\alpha^t_I}{\eta}\mathcal{B}_{\psi}({\bm{\theta}}^{t+1}_I, {\bm{\theta}}^t_I) \\
    \leq & \frac{1}{\eta}   
        \mathcal{B}_{\psi}(\sigma_i(I), {\bm{\theta}}^1_I) +  \sum^T_{t=1} \eta \frac{\Vert {\bm{r}}^{t}_I -\bm{r}^{t-1}_I \Vert^2_2}{2(1+\alpha^t_I)} + \sum^T_{t=1} \eta \frac{\alpha^t_I\Vert {\bm{R}}^{t+1}_I -{\bm{R}}^t_I \Vert^2_2}{2},
    \end{aligned}
    \end{equation}
    where {the second inequality comes from that $\forall b,c,\rho>0, bc \leq  b^2/(2\rho) + \rho c^2/2$ ($b = \Vert {\bm{r}}^{t}_I -\bm{r}^{t-1}_I \Vert_2$, $c =  \Vert {\bm{\theta}}^{t+1}_I - \hat{\bm{\theta}}^t_I \Vert_2$, and $\rho = (1+\alpha^t_I)/\eta$), the third inequality is from $\mathcal{B}_{\psi}({\bm{\theta}}^{t+1}_I, \hat{\bm{\theta}}^t_I) = \Vert {\bm{\theta}}^{t+1}_I-\hat{\bm{\theta}}^t_I \Vert^2_2/2$ (which is from the fact that in PCFR$^+$ variants, $\psi(\cdot) = \Vert \cdot \Vert^2_2/2$), as well as the last line comes from the facts that $\mathcal{B}_{\psi}({\bm{\theta}}^{t+1}_I, {\bm{\theta}}^t_I) = \Vert {\bm{\theta}}^{t+1}_I - {\bm{\theta}}^t_I \Vert^2_2/2$ and $\eta \bm{R}^t_I = \bm{\theta}^t_I$ (see more details in Section D of \citet{farina2021faster})}. For the term $\mathcal{B}_{\psi}(\sigma_i(I), {\bm{\theta}}^1_I)$ and , we have
    \begin{equation}\label{eq:sec4-temp-15-1}
    %
    \begin{aligned}
    \mathcal{B}_{\psi}(\sigma_i(I), {\bm{\theta}}^1_I) = \frac{\Vert \sigma_i(I) - {\bm{\theta}}^1_I \Vert^2_2}{2} = \frac{\Vert \sigma_i(I) \Vert^2_2}{2} \leq \frac{1}{2},
    \end{aligned}
    \end{equation}
    {where the second equality comes from the definition of ${\bm{\theta}}^1_I$, \ie, ${\bm{\theta}}^1_I = \bm{0}$, and the inequality is from $\Vert \sigma_i(I) \Vert_1 \leq 1, \forall \sigma_i(I) \in \Delta^{|A(I)|}$.} By substituting Eq. (\ref{eq:sec4-temp-15-1}) into Eq. (\ref{eq:sec4-temp-15}), we have
        \begin{equation}\label{eq:sec4-temp-16}
    %
    \begin{aligned}
    & R^{T}(I) 
    \leq  \frac{1}{2\eta}   
         +  \sum^T_{t=1} \eta \frac{\Vert {\bm{r}}^{t}_I -\bm{r}^{t-1}_I \Vert^2_2}{2(1+\alpha^t_I)} + \sum^T_{t=1} \eta \frac{\alpha^t_I\Vert {\bm{R}}^{t+1}_I -{\bm{R}}^t_I \Vert^2_2}{2}.
    \end{aligned}
    \end{equation}
    Obviously, by setting 
    \begin{equation}\label{eq:sec4-temp-17}
    %
    \begin{aligned}
    \frac{1}{\eta} = \sqrt{\sum^T_{t=1} \left(\frac{\Vert {\bm{r}}^{t}_I -\bm{r}^{t-1}_I \Vert^2_2}{1+\alpha^t_I} + \alpha^t_I\Vert {\bm{R}}^{t+1}_I -{\bm{R}}^t_I \Vert^2_2\right)} ,
    \end{aligned}
    \end{equation}
    we have that the lower bound of the upper bound of $R^T(I)$ is
    \begin{equation}\label{eq:sec4-temp-18}
    %
    \begin{aligned}
    \sqrt{{\sum^T_{t=1}\left(\frac{\Vert {\bm{r}}^{t}_I -\bm{r}^{t-1}_I \Vert^2_2}{1+\alpha^t_I} + \alpha^t_I \Vert {\bm{R}}^{t+1}_I -{\bm{R}}^t_I \Vert^2_2\right)}}.
    \end{aligned}
    \end{equation}
    It completes the proof.
\end{proof}

\section{An Alternative Regret Upper Bound of APCFR$^+$}\label{sec:Another Regret Bound of APCFR}

\begin{theorem}\label{thm:regret bound of P2PCFR-2}
Assume that $T$ iterations of APCFR$^+$ with any $\alpha^t_I \geq 0$ are conducted. Then the counterfactual regret at any infoset $I \in \mathcal{I}$ is bound by
\[
R^{T}(I) \leq  \sqrt{\sum^T_{t=1} \Vert {\bm{r}}^{t}_I - \frac{1}{1+\alpha^t_I} \bm{r}^{t-1}_I \Vert^2_2}.
\]
\end{theorem}

Obviously, by setting $\alpha \to \infty$, we derive the original bound of CFR$^+$ in its original version. Moreover, the bound in Theorem \ref{thm:regret bound of P2PCFR-2} can be combined with the bound in Theorem \ref{thm:regret bound of P2PCFR} to provide a new regret upper bound for APCFR$^+$. Specifically, the regret of APCFR$^+$ is smaller than the minimum of the bounds presented in Theorem \ref{thm:regret bound of P2PCFR} and Theorem \ref{thm:regret bound of P2PCFR-2}:
\[
R^{T}(I) \leq  \min \left( \sqrt{{\sum^T_{t=1}\left(\frac{\Vert {\bm{r}}^{t}_I -\bm{r}^{t-1}_I \Vert^2_2}{1+\alpha^t_I} + \alpha^t_I \Vert {\bm{R}}^{t+1}_I -{\bm{R}}^t_I \Vert^2_2\right)}},\ \sqrt{\sum^T_{t=1} \Vert {\bm{r}}^{t}_I - \frac{1}{1+\alpha^t_I} \bm{r}^{t-1}_I \Vert^2_2} \right).
\]

We employ Theorem \ref{thm:regret bound of P2PCFR} rather than Theorem \ref{thm:regret bound of P2PCFR-2} in the main text because the regret upper bound presented in Theorem \ref{thm:regret bound of P2PCFR-2} is typically significantly larger than that in Theorem \ref{thm:regret bound of P2PCFR}. In our experiments (as shown in Figures \ref{fig:dynamics_regret_bound}, \ref{fig:dynamics_regret_bound_subgames}, \ref{fig:dynamics_regret_bound2}, and \ref{fig:dynamics_regret_bound2_subgames}), we observe that the value of $\sum^T_{t=1} \Vert {\bm{r}}^{t}_I - \frac{1}{1+\alpha^t_I} \bm{r}^{t-1}_I \Vert^2_2$ consistently increases over time. In contrast, the value of $\sum^T_{t=1}\left(\frac{\Vert {\bm{r}}^{t}_I -\bm{r}^{t-1}_I \Vert^2_2}{1+\alpha^t_I} + \alpha^t_I \Vert {\bm{R}}^{t+1}_I -{\bm{R}}^t_I \Vert^2_2\right)$ tends to stabilize, exhibiting a flattening trend. We hypothesize that this phenomenon arises from the fact that even when ${\bm{r}}^{t}_I$ and $\bm{r}^{t-1}_I$ are close, the value of $\Vert {\bm{r}}^{t}_I - \frac{1}{1+\alpha^t_I} \bm{r}^{t-1}_I \Vert^2_2$ remains extremely large.

\begin{proof}
    Firstly, from Eq. (\ref{eq:sec4-temp-9}), we get
    \begin{equation}\label{eq:sec-B-temp-0}
    \begin{aligned}
        & \frac{1}{1+\alpha^t_I}\langle  -\bm{r}^{t-1}_I, \hat{\bm{\theta}}^t_I - {\bm{\theta}}^{t+1}_I \rangle  \leq  \frac{1}{\eta} \left( \mathcal{B}_{\psi}({\bm{\theta}}^{t+1}_I, {\bm{\theta}}^t_I) - \mathcal{B}_{\psi}({\bm{\theta}}^{t+1}_I, \hat{\bm{\theta}}^t_I)  - \mathcal{B}_{\psi}(\hat{\bm{\theta}}^t_I, {\bm{\theta}}^t_I)  \right).
    \end{aligned}
    \end{equation}
    By summing up Eq. (\ref{eq:sec:prf:lem:update-rule-SExRM+-inequality-0}) with Eq. (\ref{eq:sec-B-temp-0}), we have
    \begin{equation}\label{eq:sec-B-temp-1}
    %
    \begin{aligned}
        & \langle  -{\bm{r}}^{t}_I, {\bm{\theta}}^{t+1}_I - \hat{\bm{\theta}}^t_I + \hat{\bm{\theta}}^t_I- \sigma_i(I) \rangle + \frac{1}{1+\alpha^t_I} \langle  -\bm{r}^{t-1}_I, \hat{\bm{\theta}}^t_I - {\bm{\theta}}^{t+1}_I \rangle \\
        \leq & \frac{1}{\eta} \left(
        \mathcal{B}_{\psi}(\sigma_i(I), {\bm{\theta}}^t_I) - \mathcal{B}_{\psi}(\sigma_i(I), {\bm{\theta}}^{t+1}_I) - \mathcal{B}_{\psi}({\bm{\theta}}^{t+1}_I, {\bm{\theta}}^t_I) \right)  +  
        \\ & \quad \quad \quad
        \frac{1}{\eta} \left(\mathcal{B}_{\psi}({\bm{\theta}}^{t+1}_I, {\bm{\theta}}^t_I)- \mathcal{B}_{\psi}({\bm{\theta}}^{t+1}_I, \hat{\bm{\theta}}^t_I) - \mathcal{B}_{\psi}({\hat{\bm{\theta}}}^{t}_I, {\bm{\theta}}^t_I)  \right) \\
        \leq & \frac{1}{\eta} \left(
        \mathcal{B}_{\psi}(\sigma_i(I), {\bm{\theta}}^t_I) - \mathcal{B}_{\psi}(\sigma_i(I), {\bm{\theta}}^{t+1}_I) 
        \right)   +  \frac{1}{\eta} \left(- \mathcal{B}_{\psi}({\bm{\theta}}^{t+1}_I, \hat{\bm{\theta}}^t_I) - \mathcal{B}_{\psi}({\hat{\bm{\theta}}}^{t}_I, {\bm{\theta}}^t_I)  \right).
    \end{aligned}
    \end{equation}
    Then, we have
    \begin{equation}\label{eq:sec-B-temp-2}
    %
    \begin{aligned}
        & \langle  -{\bm{r}}^{t}_I, \hat{\bm{\theta}}^t_I- \sigma_i(I) \rangle  \\
        \leq & \langle  {\bm{r}}^{t}_I, {\bm{\theta}}^{t+1}_I - \hat{\bm{\theta}}^t_I \rangle + \frac{1}{1+\alpha^t_I} \langle  \bm{r}^{t-1}_I, \hat{\bm{\theta}}^t_I - {\bm{\theta}}^{t+1}_I \rangle + 
        \frac{1}{\eta} \left( \mathcal{B}_{\psi}(\sigma_i(I), {\bm{\theta}}^t_I) - \mathcal{B}_{\psi}(\sigma_i(I), {\bm{\theta}}^{t+1}_I) \right) +  
        \\ & \quad \quad \quad
        \frac{1}{\eta} \left(- \mathcal{B}_{\psi}({\bm{\theta}}^{t+1}_I, \hat{\bm{\theta}}^t_I) - \mathcal{B}_{\psi}({\hat{\bm{\theta}}}^{t}_I, {\bm{\theta}}^t_I)  \right).
    \end{aligned}
    \end{equation}
    Combining Eq. (\ref{eq:sec4-temp-10-1}), (\ref{eq:sec4-temp-13}) and (\ref{eq:sec-B-temp-2}), we have
    \begin{equation}\label{eq:sec-B-temp-3}
%
    \begin{aligned}
         R^{T}(I) 
        \leq & \frac{1}{\eta} \mathcal{B}_{\psi}(\sigma_i(I), {\bm{\theta}}^1_I) + 
        \sum^T_{t=1} \langle  {\bm{r}}^{t}_I, {\bm{\theta}}^{t+1}_I - \hat{\bm{\theta}}^t_I \rangle + 
        \\ & \quad \quad \quad
        \sum^T_{t=1} \frac{1}{1+\alpha^t_I} \langle  \bm{r}^{t-1}_I, \hat{\bm{\theta}}^t_I - {\bm{\theta}}^{t+1}_I \rangle   +  \sum^T_{t=1} \frac{1}{\eta} \left(- \mathcal{B}_{\psi}({\bm{\theta}}^{t+1}_I, \hat{\bm{\theta}}^t_I) - \mathcal{B}_{\psi}({\hat{\bm{\theta}}}^{t}_I, {\bm{\theta}}^t_I)  \right) \\
        \leq & \frac{1}{\eta} \mathcal{B}_{\psi}(\sigma_i(I), {\bm{\theta}}^1_I) + \sum^T_{t=1} \langle  {\bm{r}}^{t}_I - \frac{1}{1+\alpha^t_I} \bm{r}^{t-1}_I , {\bm{\theta}}^{t+1}_I - \hat{\bm{\theta}}^t_I \rangle    + 
        \sum^T_{t=1} \frac{1}{\eta} \left(- \mathcal{B}_{\psi}({\bm{\theta}}^{t+1}_I, \hat{\bm{\theta}}^t_I) - \mathcal{B}_{\psi}({\hat{\bm{\theta}}}^{t}_I, {\bm{\theta}}^t_I)  \right) \\
        \leq & \frac{1}{\eta} \mathcal{B}_{\psi}(\sigma_i(I), {\bm{\theta}}^1_I) + \sum^T_{t=1}\Vert {\bm{r}}^{t}_I - \frac{1}{1+\alpha^t_I} \bm{r}^{t-1}_I \Vert_2 \Vert {\bm{\theta}}^{t+1}_I - \hat{\bm{\theta}}^t_I \Vert_2     +  
        \sum^T_{t=1} \frac{1}{\eta} \left(- \mathcal{B}_{\psi}({\bm{\theta}}^{t+1}_I, \hat{\bm{\theta}}^t_I) - \mathcal{B}_{\psi}({\hat{\bm{\theta}}}^{t}_I, {\bm{\theta}}^t_I)  \right) \\
        \leq & \frac{1}{\eta} \mathcal{B}_{\psi}(\sigma_i(I), {\bm{\theta}}^1_I) + \sum^T_{t=1} \eta\frac{\Vert {\bm{r}}^{t}_I - \frac{1}{1+\alpha^t_I} \bm{r}^{t-1}_I \Vert^2_2}{2}  + \sum^T_{t=1} \frac{\Vert {\bm{\theta}}^{t+1}_I - \hat{\bm{\theta}}^t_I \Vert_2}{2\eta}     +  
        \sum^T_{t=1} \frac{1}{\eta} \left(- \mathcal{B}_{\psi}({\bm{\theta}}^{t+1}_I, \hat{\bm{\theta}}^t_I) - \mathcal{B}_{\psi}({\hat{\bm{\theta}}}^{t}_I, {\bm{\theta}}^t_I)  \right) \\
        \leq & \frac{1}{\eta} \mathcal{B}_{\psi}(\sigma_i(I), {\bm{\theta}}^1_I) + \sum^T_{t=1} \eta\frac{\Vert {\bm{r}}^{t}_I - \frac{1}{1+\alpha^t_I} \bm{r}^{t-1}_I \Vert^2_2}{2} ,
    \end{aligned}
    \end{equation}
    where the third inequality comes from that $\forall b,c,\rho>0, bc \leq  b^2/(2\rho) + \rho c^2/2$ (in this case, $b = \Vert {\bm{r}}^{t}_I - \frac{1}{1+\alpha^t_I} \bm{r}^{t-1}_I \Vert_2$, $c =  \Vert {\bm{\theta}}^{t+1}_I - \hat{\bm{\theta}}^t_I \Vert_2$, and $\rho = 1/\eta$), and the last inequality is from $\mathcal{B}_{\psi}({\bm{\theta}}^{t+1}_I, \hat{\bm{\theta}}^t_I) = \Vert {\bm{\theta}}^{t+1}_I-\hat{\bm{\theta}}^t_I \Vert^2_2/2$. 
    Then, from Eq. (\ref{eq:sec4-temp-15-1}) ($\mathcal{B}_{\psi}(\sigma_i(I), {\bm{\theta}}^1_I) \leq \frac{1}{2}$), we get
    \begin{equation}\label{eq:sec-B-temp-3}
%
    \begin{aligned}
        R^{T}(I) 
        \leq & \frac{1}{2\eta} + \sum^T_{t=1} \eta\frac{\Vert {\bm{r}}^{t}_I - \frac{1}{1+\alpha^t_I} \bm{r}^{t-1}_I \Vert^2_2}{2}.
    \end{aligned}
    \end{equation}
    Obviously, by setting 
    \begin{equation}\label{eq:sec-B-temp-4}
%
    \begin{aligned}
    \frac{1}{\eta} = \sqrt{\sum^T_{t=1} \Vert {\bm{r}}^{t}_I - \frac{1}{1+\alpha^t_I} \bm{r}^{t-1}_I \Vert^2_2} ,
    \end{aligned}
    \end{equation}
    we have that the lower bound of the upper bound of $R^T(I)$ is
    \begin{equation}\label{eq:sec4-temp-18}
%
    \begin{aligned}
    \sqrt{\sum^T_{t=1} \Vert {\bm{r}}^{t}_I - \frac{1}{1+\alpha^t_I} \bm{r}^{t-1}_I \Vert^2_2}.
    \end{aligned}
    \end{equation}
    It completes the proof.
\end{proof}

{
\section{Details of APDCFR$^+$}\label{sec:details of APDCFR}


\noindent Next, we introduce APDCFR$^+$. At each iteration $t$, the strategy is updated at each infoset $I \in \mathcal{I}$ as follows:
\begin{equation}\label{eq:update-rule-APDCFR+}
\begin{aligned}
    & \hat{\bm{R}}^{t}_I = \left[ \frac{\lambda t^\beta}{\kappa+t^\beta} \bm{R}^{t}_I + \frac{1}{1 + \alpha^t_I} \bm{r}^{t-1}_I \right]^+,\ \bm{R}^{t+1}_I = \left[  \bm{R}^{t}_I + \frac{\lambda t^\beta}{\kappa+t^\beta} \bm{r}^{t}_I \right]^+, \\
    & \sigma^{t}_i(I) = \frac{\left[ \hat{\bm{R}}^{t}_I \right]^+}{\| \left[ \hat{\bm{R}}^{t}_I \right]^+ \|_1} = \frac{\hat{\bm{R}}^{t}_I}{\| \hat{\bm{R}}^{t}_I \|_1},\  \alpha^t_I = \min \left( \sqrt{ \frac{\sum_{\tau=1}^{t-1} \| \bm{r}^{\tau}_I - \bm{r}^{\tau-1}_I \|^2_2}{ \sum_{\tau=1}^{t-1} \| \bm{R}^{\tau+1}_I - \bm{R}^{\tau}_I \|^2_2} }, 9 \right),
\end{aligned}
\end{equation}
where $i = P(I)$, ${\bm{R}}^{1}_I = \bm{0}$, and $\bm{r}^{0}_I = \bm{0}$. In updating $\alpha^t_I$, we opt for $\alpha_{max} = 9$ instead of $5$ as our observations suggest that $9$ yields superior performance within HUNL Subgames. Furthermore, we utilize the expression $\frac{\lambda t^\beta}{\kappa + t^\beta}$ for discounting accumulated regrets, in contrast to the $\frac{t^\beta}{1+t^\beta}$ used in DCFR. By setting different values for $\lambda$ and $\kappa$, we achieve a more aggressive discounting strategy than that in the original DCFR. Specifically, within this study, we assign $\lambda = 20$ and $\kappa = 500$. This configuration implies that for $\bm{R}^{t+1}_I$, the impact of $\bm{r}^{t}_I$ near the initial stages—where $t$ approaches $1$—is relatively extremely smaller compared to employing $\frac{t^\beta}{1+t^\beta}$. Formally, by setting $\lambda = 20$ and $\kappa = 500$, the range of the values of $\frac{\lambda t^\beta}{\kappa + t^\beta}$ is $[\frac{20}{501}, 20]$, while the range of the values of $\frac{t^\beta}{1 + t^\beta}$ is $[\frac{1}{2}, 1]$. APDCFR$^+$ uses the following weighted averaging strategy:
\begin{equation}\label{eq:update-rule-APDCFR+-temp}
\begin{aligned}
\bar{\sigma}^t = \left(\frac{t-1}{t}\right)^{2.5} \bar{\sigma}^{t-1} + \sigma^t.
\end{aligned}
\end{equation}
Notably, the parameters $\alpha_{\max}=9$, $\lambda=20$, $\kappa=500$, and $\beta=1.5$ are selected from the ranges $[5,7,9]$, $[10,20,50,100]$, $[100,500,1000]$, and $[1.0, 1.5, 2.0, 2.5]$, respectively. The parameter tuning of APDCFR$^+$ is conducted exclusively on HUNL subgames to identify the optimal configuration.

}

\section{Additional Experiments}\label{sec:Additional Experiments}


\begin{table}[t]  
\centering %
\begin{tabular*}{1.0\textwidth}{@{\extracolsep{\fill}}c|rrrrr} 
\hline  
\textbf{Game} & \textbf{\#Histories} & \textbf{\#Infosets} & \textbf{\#Terminal histories} & \textbf{\#Depth} & \textbf{\#Max size of infosets} \\
\hline  
Kuhn Poker & 58 & 12 & 30 & 6 & 2 \\
Leduc Poker & 9,457 & 936 & 5,520 & 12 & 5 \\
Battleship (3,2,3) & 732,607 & 81,027 & 552,132 & 9 & 7 \\
Battleship (4,3,2) & 5,462,407 & 58,159 & 4,966,176 & 7 & 17 \\
Liar's Dice (4) & 8,181 & 1,024 & 4,080 & 12 & 4 \\
Liar's Dice (5) & 51,181 & 5,120 & 25,575 & 14 & 5 \\
Goofspiel (4) & 1,077 & 162 & 576 & 7 & 14 \\
Goofspiel (5) & 26,931 & 2,124 & 14,400 & 9 & 46 \\
Subgame 3 & 398,112,843 & 69,184 & 261,126,360 & 10 & 1,980 \\
Subgame 4 & 244,005,483 & 43,240 & 158,388,120 & 8 & 1,980 \\
Leduc Poker (5) & 55,361 & 2,760 & 32,760 & 12 & 9 \\
Leduc Poker (9) & 371,809 & 9,288 & 221,544 & 12 & 17 \\
Leduc Poker (13) & 1,179,777 & 19,656 & 704,600 & 12 & 25 \\
\hline  
\end{tabular*}  
\caption{Sizes of the games.}  
\label{tab:Number of Information Sets of Games}  
\end{table}

\noindent \textbf{Sizes of the Games.} Before introducing our additional experiments, we present the sizes of the games used in our experiments, as shown in \Cref{tab:Number of Information Sets of Games}. In this table, \#Histories denotes the number of histories in the game tree, while \#Infosets represents the number of infosets in the game tree. \#Terminal histories indicates the number of terminal histories in the game tree. \#Depth defines the depth of the game tree, which is the maximum number of actions in a single history. Lastly, \#Max size of infosets indicates the maximum number of histories that belong to the same infoset.

%

\begin{table*}[t]
\centering
    \centering 
    
    \begin{tabular*}{1.0\textwidth}{@{\extracolsep{\fill}}c|rrr}
        \hline
           \textbf{Game}       & PCFR$^+$ & APCFR$^+$ & SAPCFR$^+$ \\
        \hline
        kuhn Poker &0.0020 &0.0033 &0.0021 \\
        Leduc Poker &0.4252 &0.6423 &0.4408 \\
        Battleship (3,2,3) &50.1909 &58.4390 &50.8602 \\
        Battleship (4,3,2) &103.2314 &116.2511 &102.4161 \\
        Liar's Dice (4) &0.4327 &0.5690 &0.4308 \\
        Liar's Dice (5) &3.8178 &4.0365 &3.7922 \\
        Goofspiel (4) &0.0290 &0.0393 &0.0279 \\
        Goofspiel (5) &0.9323 &1.9760 &0.9430 \\
        Subgame 3 &278.7428 &297.3784 &276.9873 \\
        Subgame 4 &245.1753 &254.4792 &249.0042 \\
        \hline
    \end{tabular*}
    \caption{Comparison of running times (in minutes).}
    \label{tab:runing time}
\end{table*}

\textbf{Running times.} We now compare the running time of our algorithms with that of PCFR$^+$, both executed with the same number of iterations (i.e., 5000). The experimental results are shown in Table \ref{tab:runing time}. We observe that the running time of APCFR$^+$ is slightly higher compared to PCFR$^+$, primarily due to the additional $\alpha^t_I$ learning process in APCFR$^+$, which increases its running time. However, we also note that the running time of SAPCFR$^+$ is nearly identical to that of PCFR$^+$, as the only difference between their implementations is a single line of code, which does not alter the computational complexity. Importantly, the computational complexity remains exactly the same, even with no change in the constant factors.

\begin{figure*}[t]
    \centering 
    \subfigure{
    \includegraphics[width=1\linewidth]{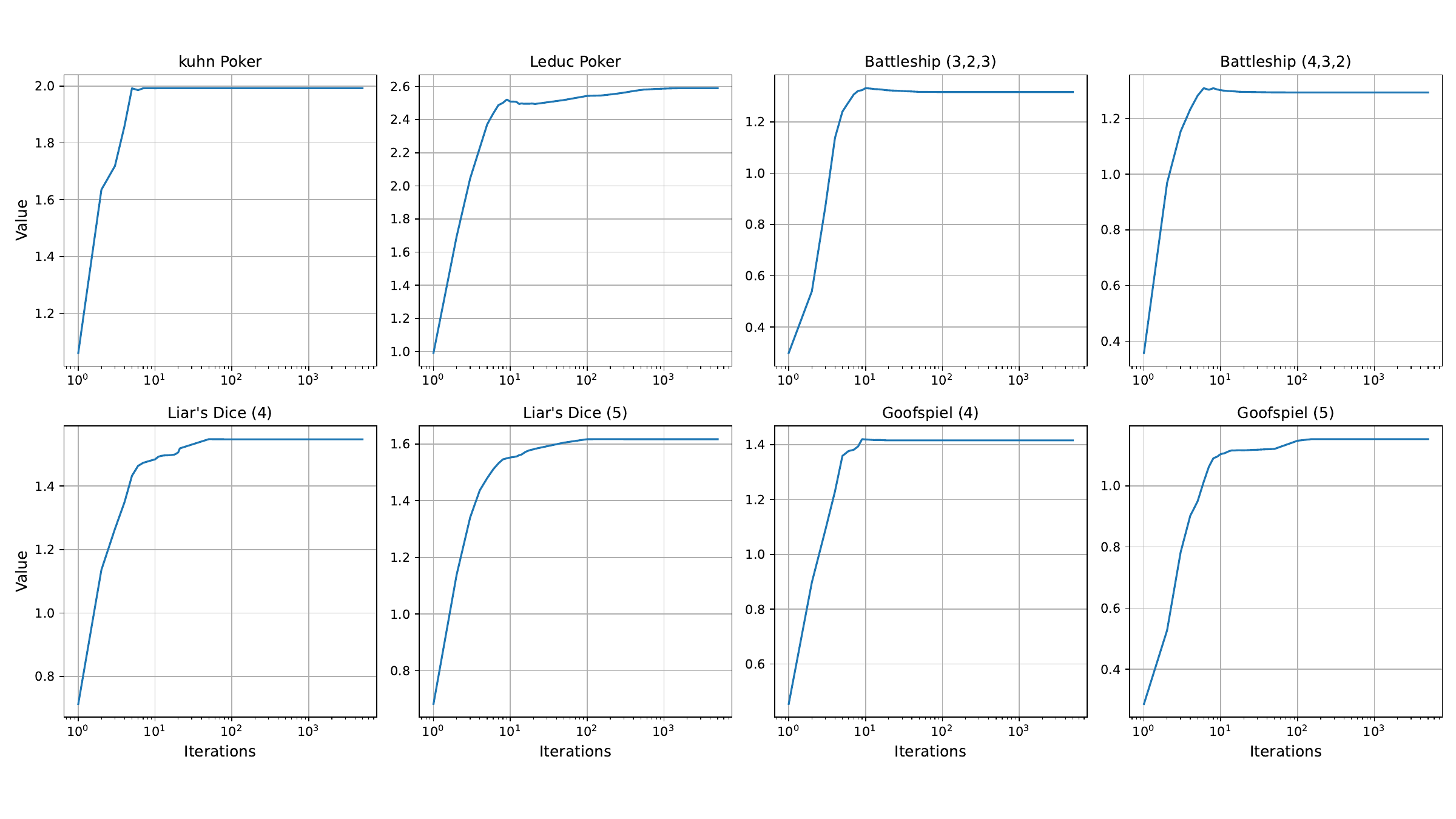}
    }
    \caption{Dynamics of $\alpha^t_I$ in standard commonly used IIG benchmarks. Note that, contrary to figures in the main text, the x-axis in this figure is on a logarithmic scale, while the y-axis is not.
}
\label{fig:dynamics_alpha}
\end{figure*}

\begin{figure}[t]
    \centering 
    \subfigure{
    \includegraphics[width=0.65\linewidth]{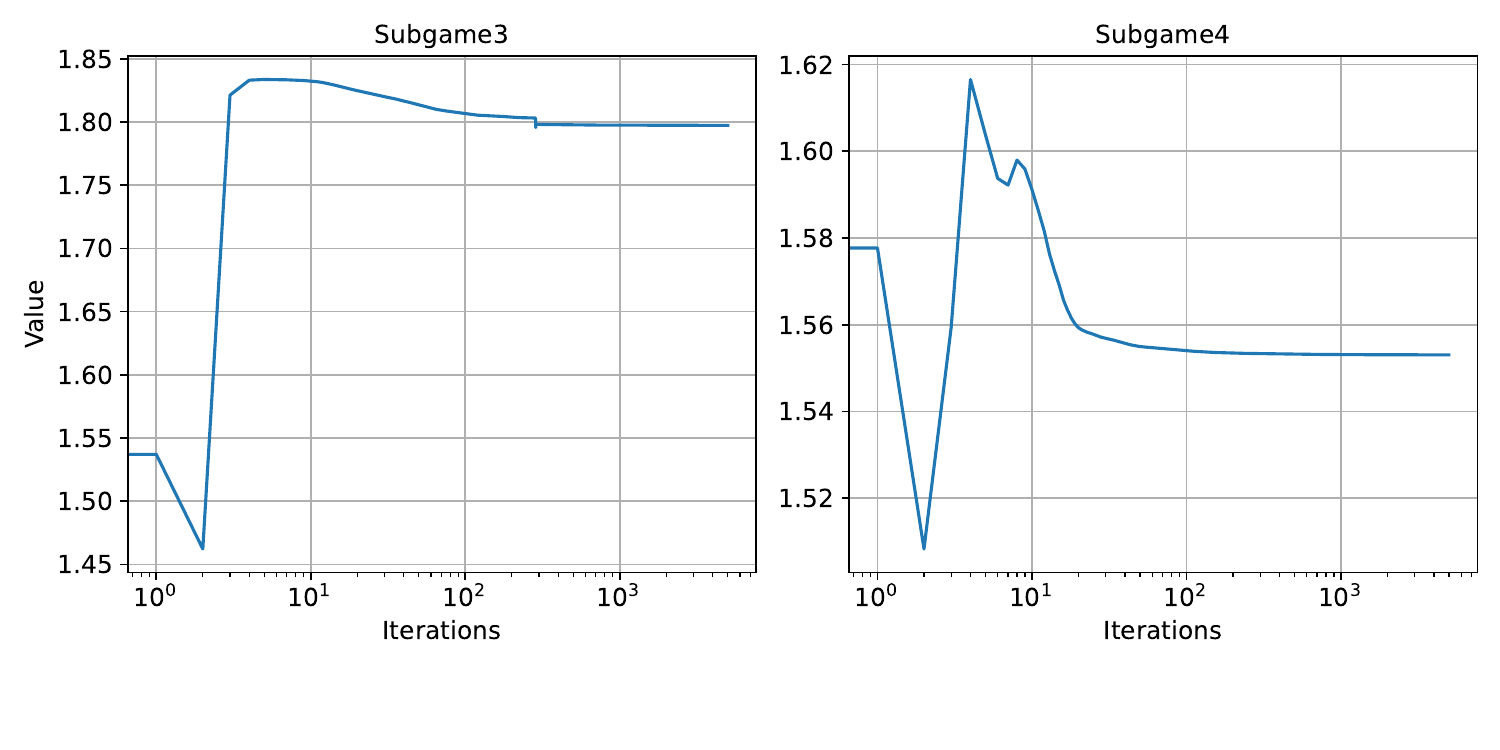}
    }
    \caption{Dynamics of $\alpha^t_I$ in HUNL Subgames.
}
\label{fig:dynamics_alpha_subgames}
\end{figure}

\textbf{Dynamics of $\alpha^t_I$ in APCFR$^+$.} We now present the dynamics of $\alpha^t_I$ in APCFR$^+$. The experimental results are shown in Figures \ref{fig:dynamics_alpha} and \ref{fig:dynamics_alpha_subgames}. Note that we output the average value of $\alpha^t_I$ across all $I \in \mathcal{I}$. Initially, $\alpha^t_I$ increases rapidly. After approximately 100 iterations, $\alpha^t_I$ stops increasing. A closer examination is that the values of $\Vert {\bm{r}}^t_I - {\bm{r}}^{t-1}_I \Vert^2_2$ and $\Vert {\bm{R}}^{t+1}_I - {\bm{R}}^t_I \Vert^2_2$ are significantly larger in the initial phase than at later stages (as observed by the results shown in below).

\begin{figure*}[t!]
    \centering 
    \subfigure{
    \includegraphics[width=1\linewidth]{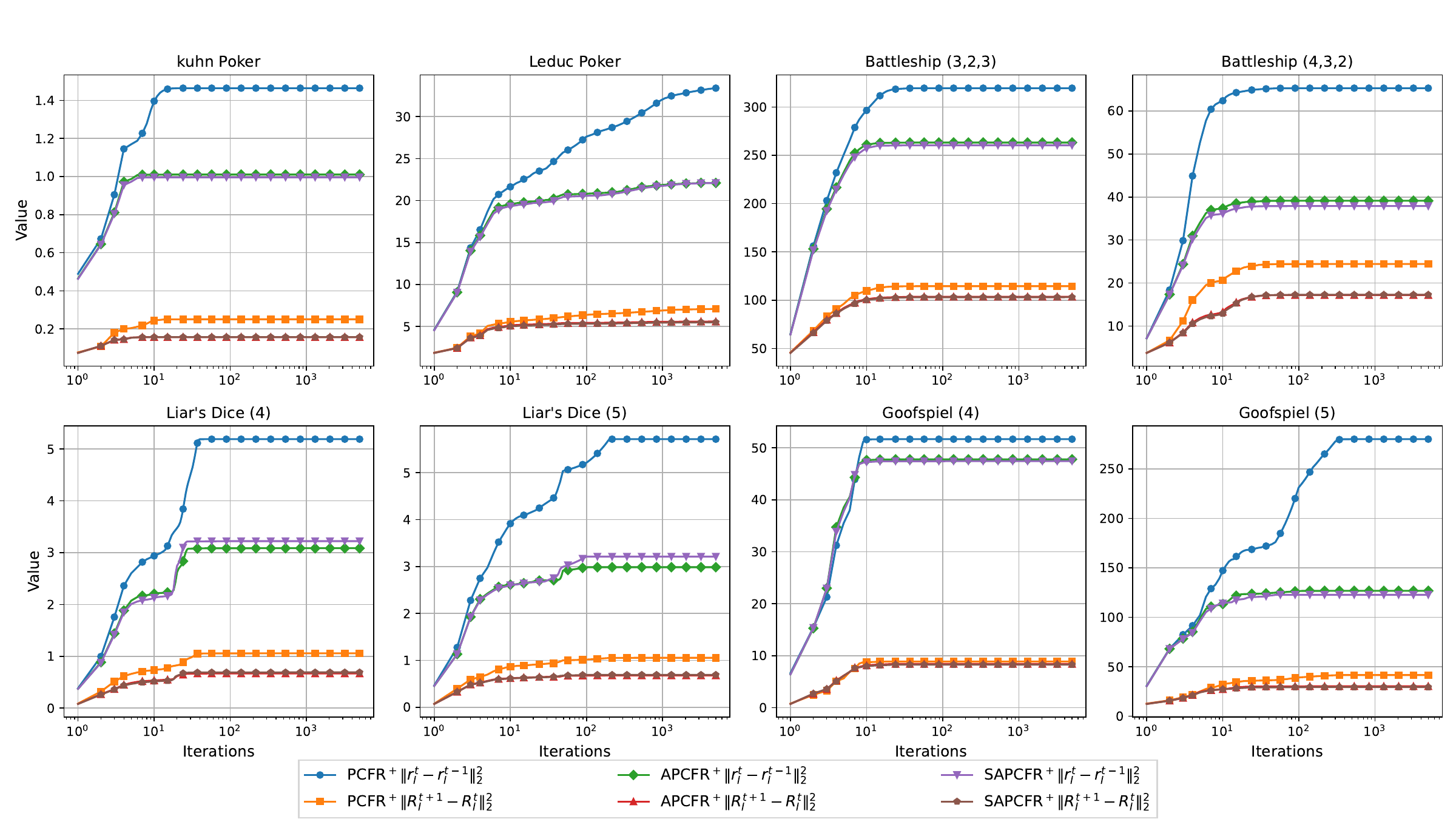}
    }
    \caption{Dynamics of $
\sum^T_{t=1} \Vert {\bm{r}}^t_I - {\bm{r}}^{t-1}_I \Vert^2_2$ and $\sum^T_{t=1} \Vert {\bm{R}}^{t+1}_I - {\bm{R}}^t_I \Vert^2_2
$ in standard commonly used IIG benchmarks.
}
\label{fig:dynamics_regret_bound_imm_gap_regret_gap_without_alpha}
\end{figure*}
\begin{figure*}[t!]
    \centering 
    \subfigure{
    \includegraphics[width=0.65\linewidth]{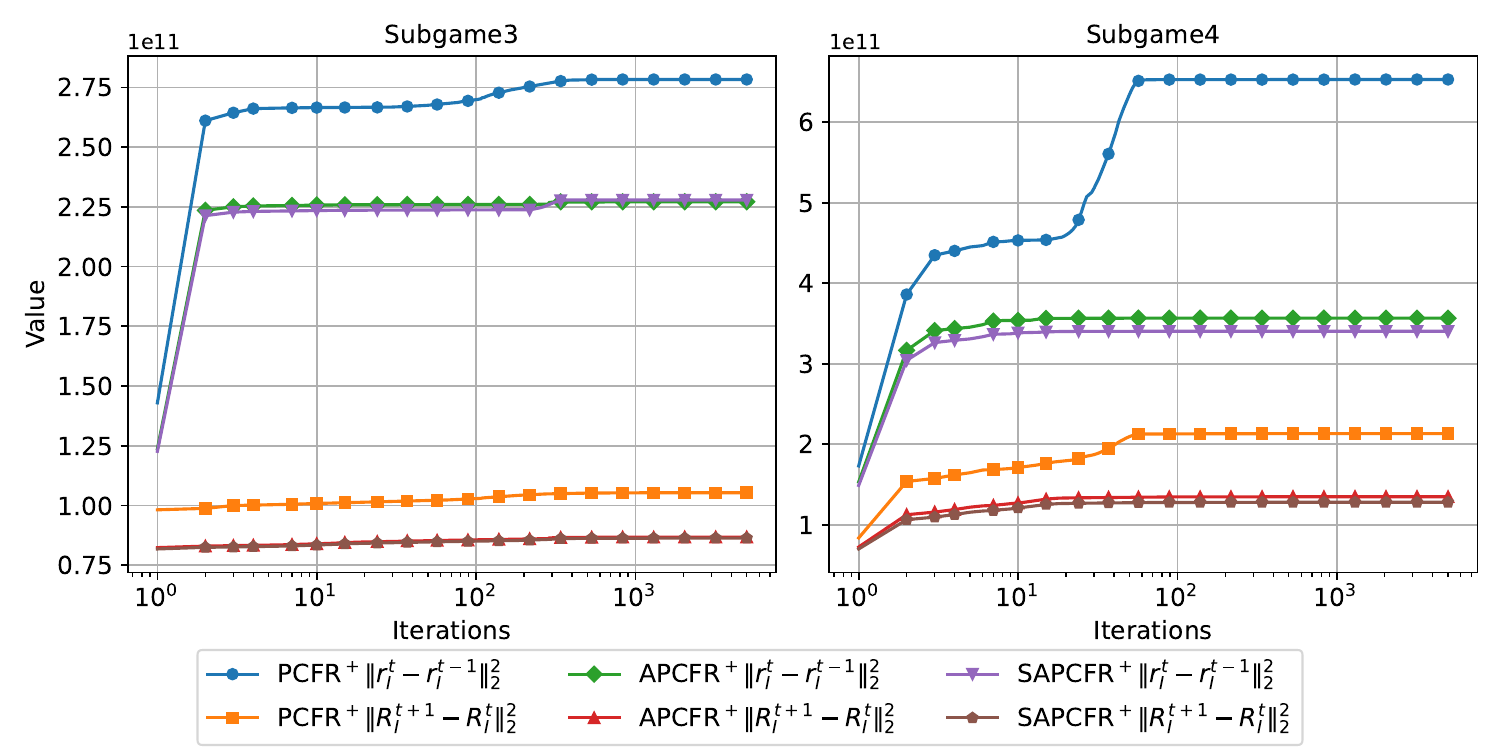}
    }
    \caption{Dynamics of $
\sum^T_{t=1} \Vert {\bm{r}}^t_I - {\bm{r}}^{t-1}_I \Vert^2_2$ and $\sum^T_{t=1} \Vert {\bm{R}}^{t+1}_I - {\bm{R}}^t_I \Vert^2_2
$ in standard commonly used IIG benchmarks.
}
\label{fig:dynamics_regret_bound_subgames_imm_gap_regret_gap_without_alpha}
\end{figure*}

{\textbf{Dynamics of 
{$
\sum_{t=1}^{T} \Vert \bm{r}^t_I - \bm{r}^{t-1}_I \Vert^2_2 \text{ and } \sum_{t=1}^{T} \Vert \bm{R}^{t+1}_I - \bm{R}^t_I \Vert^2_2
$}.}
Experimental results are shown in Figures \ref{fig:dynamics_regret_bound_imm_gap_regret_gap_without_alpha} and \ref{fig:dynamics_regret_bound_subgames_imm_gap_regret_gap_without_alpha}. We show the cumulative values across all infosets.
We observe that 
{$
\sum_{t=1}^{T} \Vert \bm{r}^t_I - \bm{r}^{t-1}_I \Vert^2_2
$}
is at least three times larger than 
{$
\sum_{t=1}^{T} \Vert \bm{R}^{t+1}_I - \bm{R}^t_I \Vert^2_2
$}
for both PCFR$^+$ and our algorithms. Intuitively, this phenomenon can be explained by the fact that the upper bound of {$\Vert {\bm{R}}^{t+1}_I - {\bm{R}}^t_I \Vert^2_2$} is only a quarter of the upper bound of {$\Vert {\bm{r}}^{t}_I - \bm{r}^{t-1}_I \Vert^2_2$}, which is also used to derive SAPCFR$^+$ (as detailed in Section \ref{subsec:SAPCFR+}).
Therefore, there must exists a sequence of $\alpha^t_I$ such that 
{$
\sum_{t=1}^{T} ( \frac{\Vert \bm{r}^t_I - \bm{r}^{t-1}_I \Vert^2_2}{1 + \alpha^t_I} + \alpha^t_I \Vert \bm{R}^{t+1}_I - \bm{R}^t_I \Vert^2_2 ) \leq \sum_{t=1}^{T} \Vert \bm{r}^t_I - \bm{r}^{t-1}_I \Vert^2_2,
$}
where the term in the right-hand side is the regret bound for PCFR$^+$.
Specifically, let $\alpha^t_I = \alpha > 0$ be a positive constant. To satisfy the inequality 
{$
\sum_{t=1}^{T} ( \frac{\Vert \bm{r}^t_I - \bm{r}^{t-1}_I \Vert^2_2}{1 + \alpha} + \alpha \Vert \bm{R}^{t+1}_I - \bm{R}^t_I \Vert^2_2 ) \leq \sum_{t=1}^{T} \Vert \bm{r}^t_I - \bm{r}^{t-1}_I \Vert^2_2,
$}
it suffices to ensure that
{$
\sum_{t=1}^{T} \alpha \Vert \bm{R}^{t+1}_I - \bm{R}^t_I \Vert^2_2 \leq \sum_{t=1}^{T} \frac{\alpha \Vert \bm{r}^t_I - \bm{r}^{t-1}_I \Vert^2_2}{1 + \alpha}
$}
(which comes from 
{$
\sum_{t=1}^{T} \alpha \Vert \bm{R}^{t+1}_I - \bm{R}^t_I \Vert^2_2 \leq \sum_{t=1}^{T} \Vert \bm{r}^t_I - \bm{r}^{t-1}_I \Vert^2_2 - \sum_{t=1}^{T} \frac{\Vert \bm{r}^t_I - \bm{r}^{t-1}_I \Vert^2_2}{1 + \alpha}
$}
), 
or equivalently,
{$
(1 + \alpha) \sum_{t=1}^{T} \Vert \bm{R}^{t+1}_I - \bm{R}^t_I \Vert^2_2 \leq \sum_{t=1}^{T} \Vert \bm{r}^t_I - \bm{r}^{t-1}_I \Vert^2_2$}.
Given that 
{$
\sum_{t=1}^{T} \Vert \bm{r}^t_I - \bm{r}^{t-1}_I \Vert^2_2
$}
is at least three times larger than
{$
\sum_{t=1}^{T} \Vert \bm{R}^{t+1}_I - \bm{R}^t_I \Vert^2_2
$}, we can choose $0 < \alpha \leq 2$ to ensure 
{$
(1 + \alpha) \sum_{t=1}^{T} \Vert \bm{R}^{t+1}_I - \bm{R}^t_I \Vert^2_2 \leq \sum_{t=1}^{T} \Vert \bm{r}^t_I - \bm{r}^{t-1}_I \Vert^2_2$} holds.
}

{Additionally, we find that both 
$
\sum_{t=1}^{T} \Vert \bm{r}^t_I - \bm{r}^{t-1}_I \Vert^2_2
$
and 
$
\sum_{t=1}^{T} \Vert \bm{R}^{t+1}_I - \bm{R}^t_I \Vert^2_2
$
decrease for our algorithms (APCFR$^+$ and SAPCFR$^+$) compared to PCFR$^+$.
We speculate that this is due to our algorithms' higher stability compared to PCFR$^+$.
Specifically, for the reduction of $\sum_{t=1}^{T} \Vert \bm{r}^t_I - \bm{r}^{t-1}_I \Vert^2_2$, when updating using the prediction, our algorithm utilizes smaller learning rates, resulting in smaller gaps between strategies indicated by different explicit accumulated regrets. Intuitively, it leads to smaller value of $\Vert \bm{r}^t_I - \bm{r}^{t-1}_I \Vert^2_2$. For the reduction of $\sum_{t=1}^{T} \Vert \bm{R}^{t+1}_I - \bm{R}^t_I \Vert^2_2$, we attribute it to that as the learning rate decreases, the gap between the strategies represented by $\bm{R}^t_I$ and $\hat{\bm{R}^t_I}$ also diminishes, which implies the $\bm{r}^t_I$ and instantaneous counterfactual regret derived from the strategy represented by $\bm{R}^t_I$ becomes closer. This leads to more stable updates from $\bm{R}^t_I$ to $\bm{R}^{t+1}_I$. }

{Furthermore, we observe that the improvement of our algorithms over PCFR$^+$ is correlated with the rate at which the inaccuracy $\Vert \bm{r}^t_I - \bm{r}^{t-1}_I \Vert^2_2$ between the predicted and observed instantaneous counterfactual regrets decreases in PCFR$^+$. When considering the sizes of the games (\Cref{tab:Number of Information Sets of Games}), this inaccuracy decreases particularly rapidly in Battleship (3,2,3) and Goofspiel (4). Specifically, the value of the accumulated inaccuracy $\sum_{t=1}^{T} \Vert \bm{r}^t_I - \bm{r}^{t-1}_I \Vert^2_2$ stops increasing after approximately 10 iterations in only Kuhn Poker, Battleship (3,2,3), and Goofspiel (4), with Kuhn Poker is significantly smaller than Battleship (3,2,3) and Goofspiel (4). According to the experimental results in \Cref{fig:PSPCFR+-with-other-algorithms}, these games are also the games where our algorithms outperform similar with PCFR$^+$.}

{We also observe that 
$
\sum_{t=1}^{T} \Vert \bm{r}^t_I - \bm{r}^{t-1}_I \Vert^2_2
$
and 
$
\sum_{t=1}^{T} \Vert \bm{R}^{t+1}_I - \bm{R}^t_I \Vert^2_2
$
exhibit a much higher growth rate during the initial iterations (i.e., when the number of iterations is less than 100) compared to later iterations (i.e., when the number of iterations exceeds 100). 
This observation aligns with the experimental results on the dynamics of $\alpha^t_I$, showing that these terms are significantly larger in the early stages of iteration than in later stages.}

\begin{figure*}[t!]
    \centering 
    \subfigure{
    \includegraphics[width=1\linewidth]{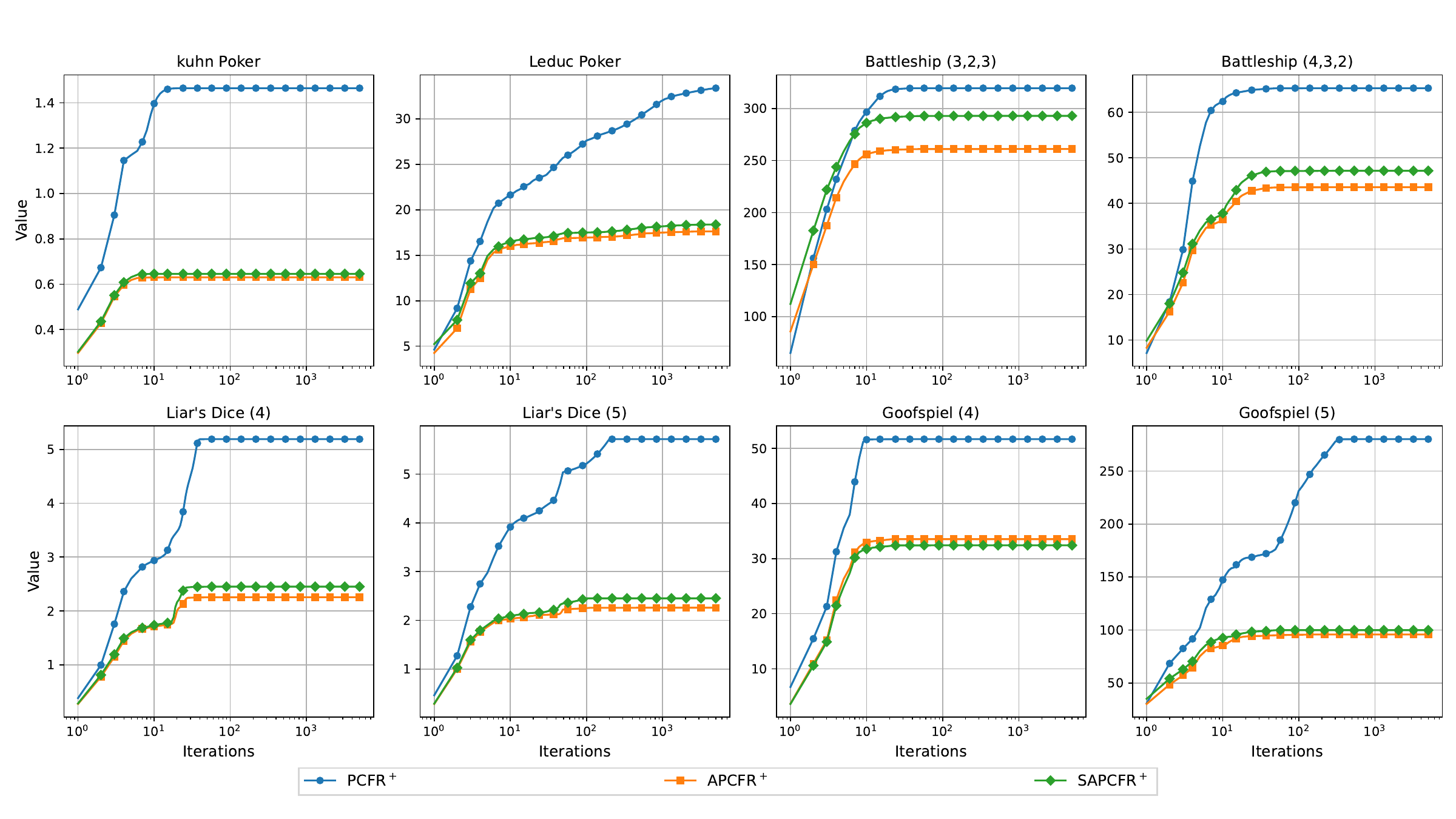}
    }
    \caption{Dynamics of $
{\sum^T_{t=1} ( \frac{\Vert {\bm{r}}^t_I - {\bm{r}}^{t-1}_I \Vert^2_2}{1 + \alpha^t_I} + \alpha^t_I \Vert {\bm{R}}^{t+1}_I - {\bm{R}}^t_I \Vert^2_2 )}
$ in standard commonly used IIG benchmarks.
}
\label{fig:dynamics_regret_bound}
\end{figure*}

\begin{figure*}[t!]
    \centering 
    \subfigure{
    \includegraphics[width=0.65\linewidth]{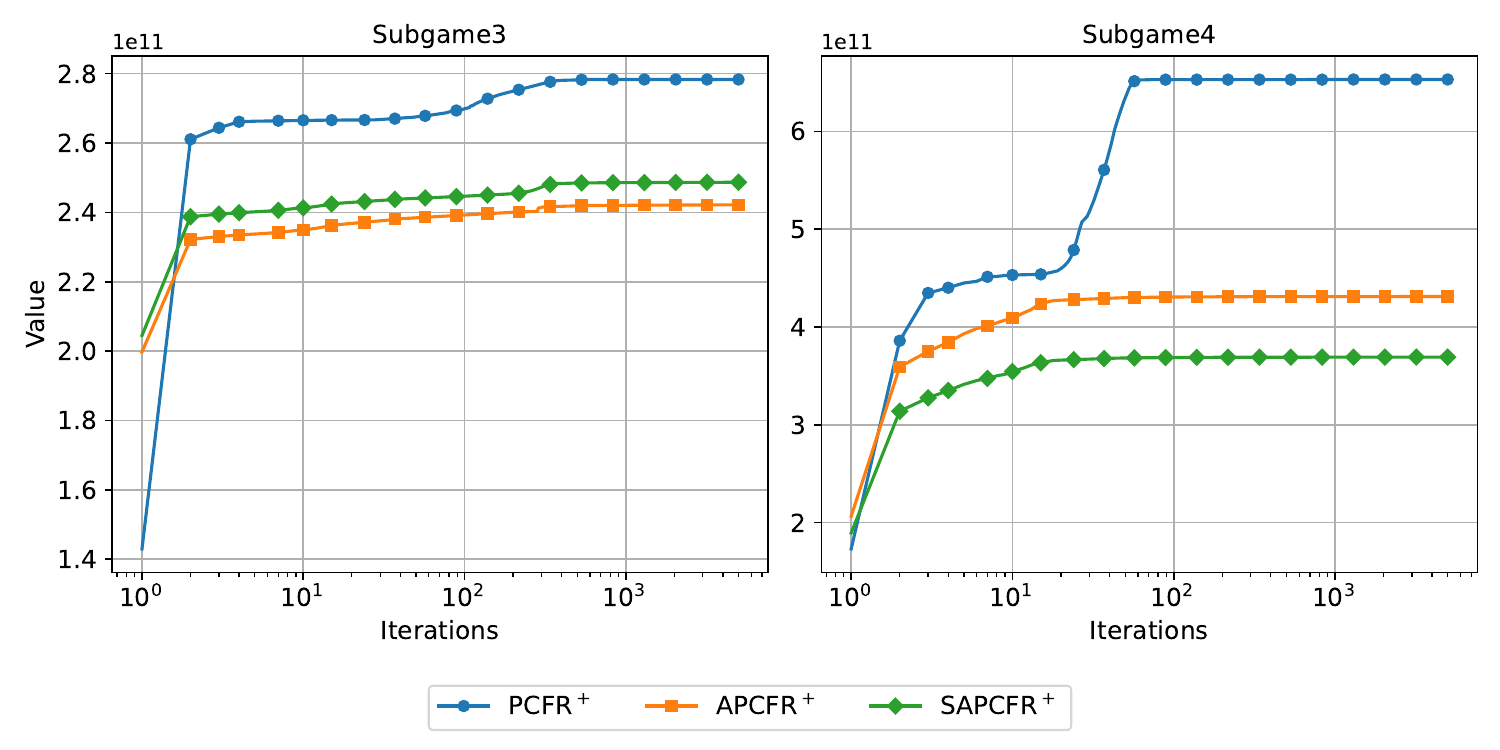}
    }
    \caption{Dynamics of $
{\sum^T_{t=1} ( \frac{\Vert {\bm{r}}^t_I - {\bm{r}}^{t-1}_I \Vert^2_2}{1 + \alpha^t_I} + \alpha^t_I \Vert {\bm{R}}^{t+1}_I - {\bm{R}}^t_I \Vert^2_2 )}
$ in HUNL Subgames.
}
\label{fig:dynamics_regret_bound_subgames}
\end{figure*}

\textbf{Dynamics of} 
$
\sum_{t=1}^{T} ( \frac{\Vert \bm{r}^t_I - \bm{r}^{t-1}_I \Vert^2_2}{1 + \alpha^t_I} + \alpha^t_I \Vert \bm{R}^{t+1}_I - \bm{R}^t_I \Vert^2_2 )
$.
The dynamics of 
$
\sum_{t=1}^{T} ( \frac{\Vert \bm{r}^t_I - \bm{r}^{t-1}_I \Vert^2_2}{1 + \alpha^t_I} + \alpha^t_I \Vert \bm{R}^{t+1}_I - \bm{R}^t_I \Vert^2_2 )
$
for our algorithms and PCFR$^+$, which is related to the upper bound of the counterfactual regret presented in Theorem \ref{thm:regret bound of P2PCFR}, are shown in Figures \ref{fig:dynamics_regret_bound} and \ref{fig:dynamics_regret_bound_subgames}.
These figures show the cumulative values across all infosets.
We observe that the values of our algorithms, APCFR$^+$ and SAPCFR$^+$, are lower than those of PCFR$^+$. 
This indicates that our algorithms achieve a lower upper bound on the counterfactual regret.
Notably, we find that for most cases, the values of this term are smaller in APCFR$^+$ than in SAPCFR$^+$, with exceptions occurring only in Goofspiel (4) and Subgame 4.
A lower upper bound on the counterfactual regret implies a faster convergence rate, which is consistent with the experimental results presented in Figure \ref{fig:PSPCFR+-with-other-algorithms} and Table \ref{tab:PSPCFR+-with-other-algorithms-HUNL Subgames}.

\begin{figure*}[t!]
    \centering 
    \subfigure{
    \includegraphics[width=1\linewidth]{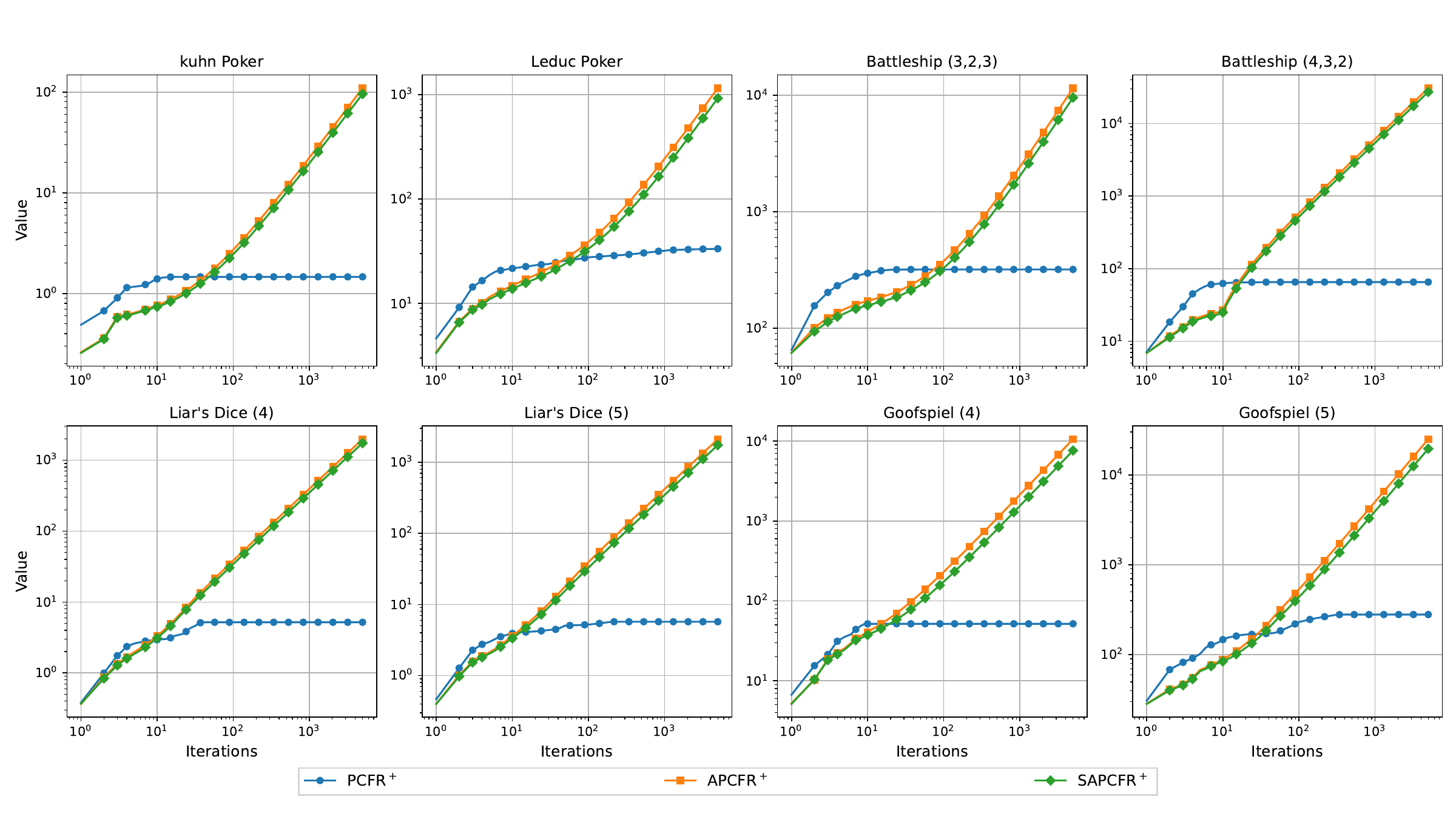}
    }
    \caption{Dynamics of $ {\sum^T_{t=1} \Vert {\bm{r}}^{t}_I - \frac{{\bm{r}}^{t-1}_I }{{1 + \alpha^t_I}}\Vert^2_2 } $ in standard commonly used IIG benchmarks.
}
\label{fig:dynamics_regret_bound2}
\end{figure*}

\begin{figure*}[t!]
    \centering 
    \subfigure{
    \includegraphics[width=0.65\linewidth]{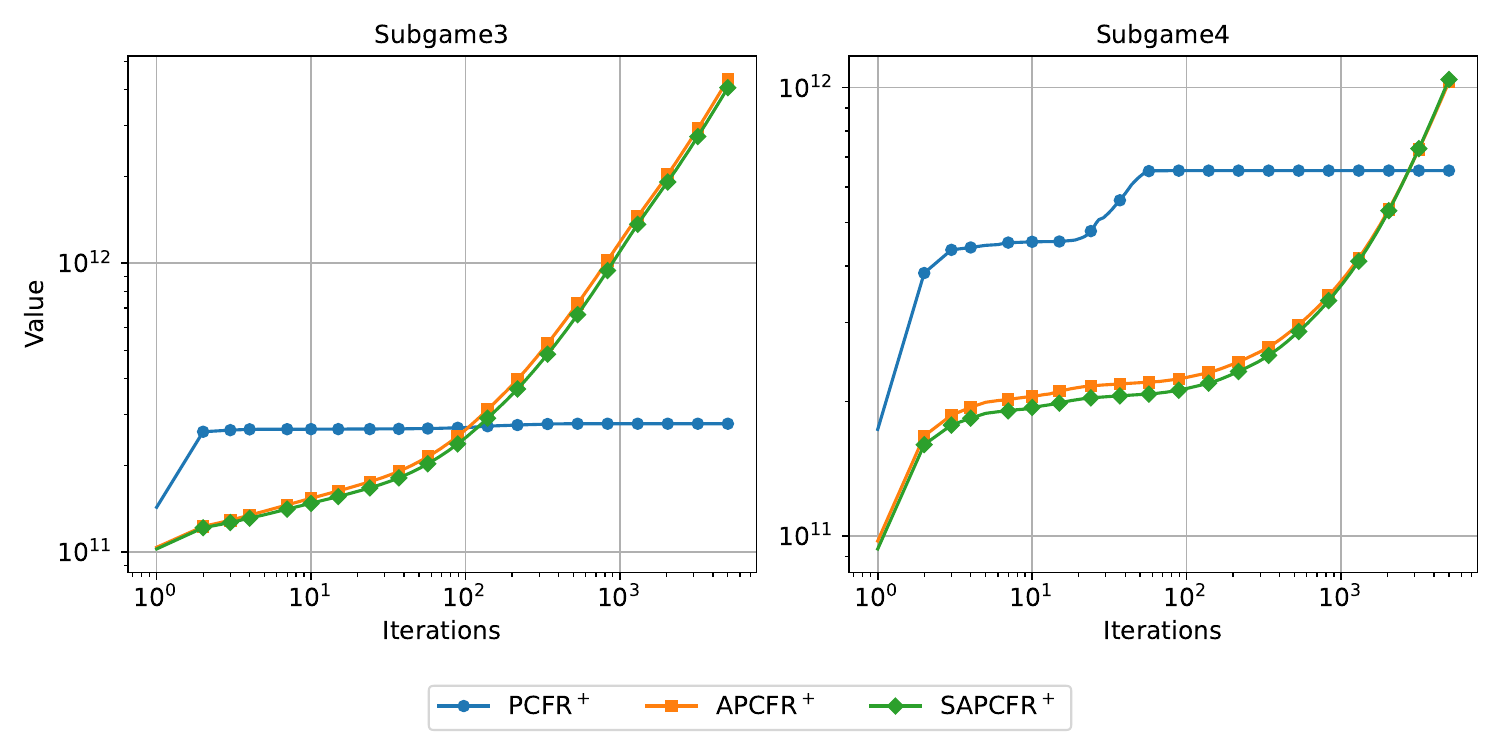}
    }
    \caption{Dynamics of $ {\sum^T_{t=1} \Vert {\bm{r}}^{t}_I - \frac{{\bm{r}}^{t-1}_I }{{1 + \alpha^t_I}}\Vert^2_2 } $ in HUNL Subgames.
}
\label{fig:dynamics_regret_bound2_subgames}
\end{figure*}

\textbf{Dynamics of $\sum^T_{t=1} \Vert {\bm{r}}^{t}_I - \frac{{\bm{r}}^{t-1}_I }{{1 + \alpha^t_I}}\Vert^2_2$.} 
The dynamics of $\sum^T_{t=1} \Vert {\bm{r}}^{t}_I - \frac{{\bm{r}}^{t-1}_I }{{1 + \alpha^t_I}}\Vert^2_2$ for both our algorithms and PCFR$^+$, which is related to the upper bound on counterfactual regret presented in Theorem \ref{thm:regret bound of P2PCFR-2}, is illustrated in Figures \ref{fig:dynamics_regret_bound2} and \ref{fig:dynamics_regret_bound2_subgames}. These figures present the cumulative values across all infosets. For PCFR$^+$, the expressions 
$\sum^T_{t=1} \Vert {\bm{r}}^{t}_I - \frac{{\bm{r}}^{t-1}_I }{{1 + \alpha^t_I}}\Vert^2_2$ 
and 
$\sum_{t=1}^{T} \left( \frac{\Vert \bm{r}^t_I - \bm{r}^{t-1}_I \Vert^2_2}{1 + \alpha^t_I} + \alpha^t_I \Vert \bm{R}^{t+1}_I - \bm{R}^t_I \Vert^2_2 \right)$
are equivalent, as $\alpha^t_I = 0$ for PCFR$^+$. By synthesizing the results from Figures \ref{fig:dynamics_regret_bound}, \ref{fig:dynamics_regret_bound_subgames}, \ref{fig:dynamics_regret_bound2}, and \ref{fig:dynamics_regret_bound2_subgames}, we observe that for our algorithms, the value of 
$\sum^T_{t=1} \Vert {\bm{r}}^{t}_I - \frac{{\bm{r}}^{t-1}_I }{{1 + \alpha^t_I}}\Vert^2_2$ 
significantly exceeds that of 
$\sum_{t=1}^{T} \left( \frac{\Vert \bm{r}^t_I - \bm{r}^{t-1}_I \Vert^2_2}{1 + \alpha^t_I} + \alpha^t_I \Vert \bm{R}^{t+1}_I - \bm{R}^t_I \Vert^2_2 \right)$ 
as the number of iterations increases. Specifically, we observe that the value of 
$\sum^T_{t=1} \Vert {\bm{r}}^{t}_I - \frac{1}{1+\alpha^t_I} \bm{r}^{t-1}_I \Vert^2_2$ 
consistently increases over time. In contrast, the value of 
$\sum^T_{t=1}\left(\frac{\Vert {\bm{r}}^{t}_I -\bm{r}^{t-1}_I \Vert^2_2}{1+\alpha^t_I} + \alpha^t_I \Vert {\bm{R}}^{t+1}_I -{\bm{R}}^t_I \Vert^2_2\right)$ 
tends to stabilize, exhibiting a flattening trend. We hypothesize that this phenomenon arises because, even when ${\bm{r}}^{t}_I$ and $\bm{r}^{t-1}_I$ are close, the value of 
$\Vert {\bm{r}}^{t}_I - \frac{1}{1+\alpha^t_I} \bm{r}^{t-1}_I \Vert^2_2$ 
remains significantly large. Consequently, we employ Theorem \ref{thm:regret bound of P2PCFR} in the main text instead of Theorem \ref{thm:regret bound of P2PCFR-2}.

\begin{figure*}[t]
    \centering 
    \subfigure{
    \includegraphics[width=1\linewidth]{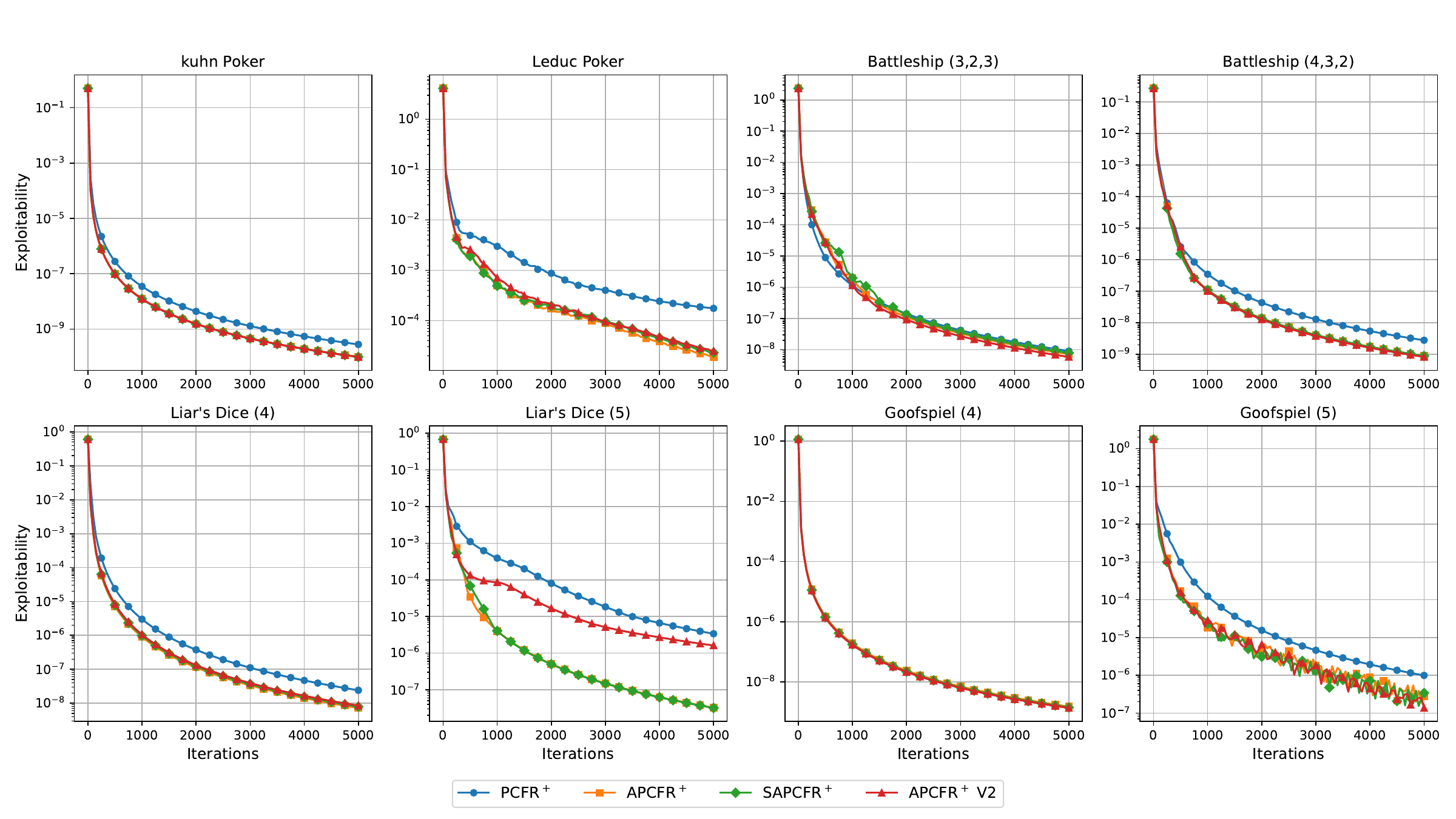}
    }
    \caption{Empirical convergence rates of PCFR$^+$, APCFR$^+$, SAPCFR$^+$, and APCFR$^+$ V2 in standard commonly used IIG benchmarks.
}
\label{fig:PSPCFR+ V2-with-other-algorithms}
\end{figure*}

\textbf{Empirical convergence rates of APCFR$^+$ with an alternative learning approach for $\alpha^t$.} In Eq (\ref{eq:Sec-4-2}), we proposed a learning approach for $\alpha^t$. Here, we evaluate an alternative learning approach for $\alpha^t$, specifically defined as $\alpha^t_I = \min\left( \sqrt{\frac{\max_{\tau \in [t-1]} \Vert \bm{r}^{\tau}_I - \bm{r}^{\tau-1}_I \Vert^2_2}{\max_{\tau \in [t-1]} \Vert \bm{R}^{\tau+1}_I - \bm{R}^{\tau}_I \Vert^2_2}}, \ \alpha_{max} \right)$. The experimental results are presented in Figure \ref{fig:PSPCFR+ V2-with-other-algorithms} and Table \ref{fig:PSPCFR+ V2-with-other-algorithms-HUNL Subgames},  where the APCFR$^+$ with new learning approach for $\alpha^t$ is denoted as ``APCFR$^+$ V2". These results demonstrate that this alternative $\alpha^t$ learning approach generally achieves performance comparable to the method outlined the original APCFR$^+$ in most games. We also observe that this alternative $\alpha^t$ learning approach slightly outperforms the original APCFR$^+$ in large games such as Battleship (3,2,3), Battleship (4,3,2), Subgame (3), and Subgame (4). A potential direction for future work is to develop more efficient algorithms for solving large IIGs, building upon the new $\alpha^t$ learning approach. However, in Liar's Dice (5), this alternative approach performs significantly worse than the method in Eq.~(\ref{eq:Sec-4-2}).

\begin{figure*}[t]
    \centering 
    \subfigure{
    \includegraphics[width=1\linewidth]{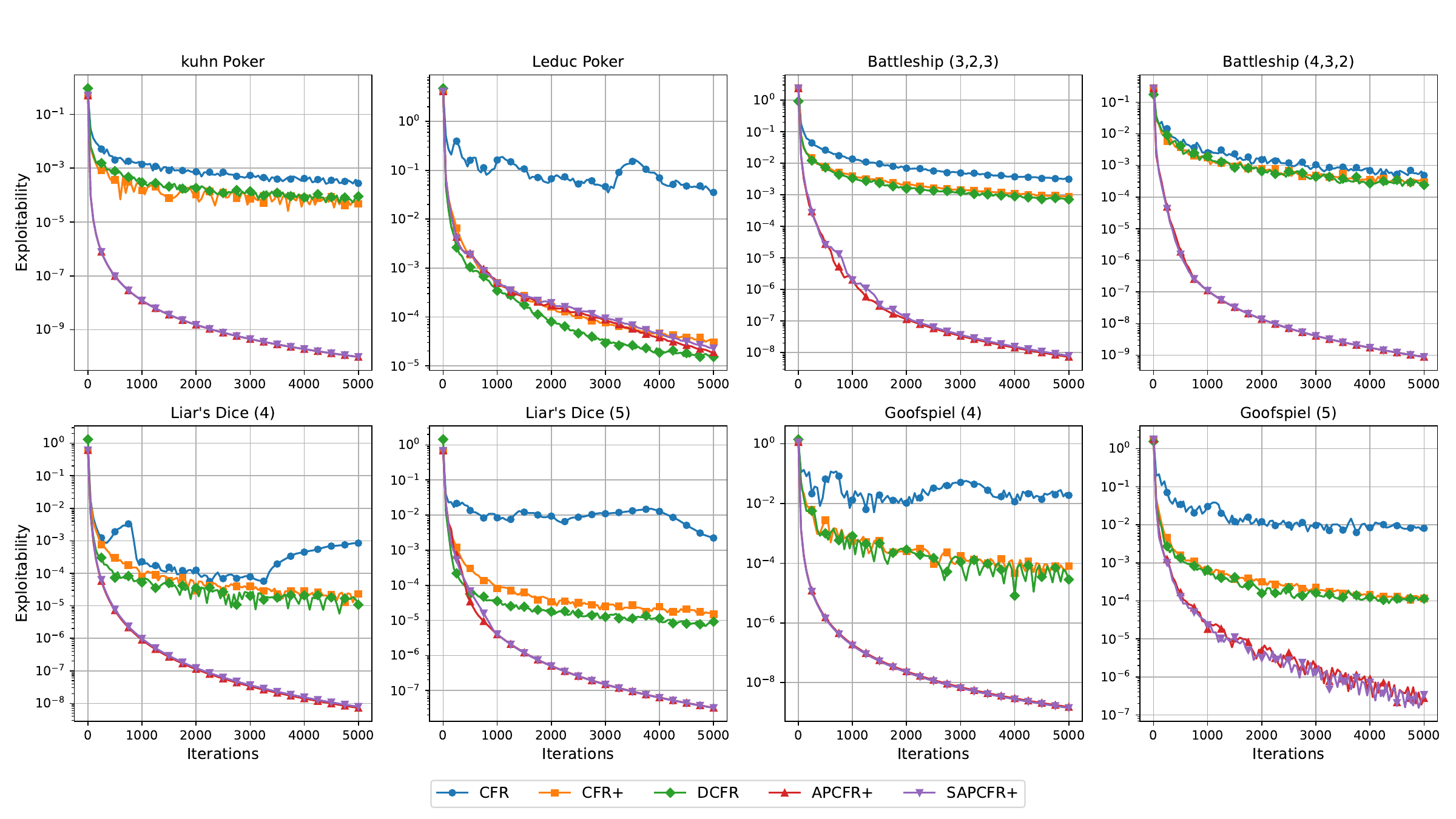}
    }
    \caption{Empirical convergence rates of CFR, CFR$^+$, DCFR APCFR$^+$, and SAPCFR$^+$ in standard commonly used IIG benchmarks.
}
\label{fig:PSPCFR+-with-other-algorithms-classical-cfr}
\end{figure*}

\textbf{Comparison with other classical CFR algorithms and the generalization of our approach.} Now, we evaluate our algorithms with other classical CFR algorithms such as CFR~\citep{zinkevich2007regret}, CFR$^+$~\citep{tammelin2014solving}, and DCFR~\citep{brown2019solving}. Initially, we conducted experiments using standard commonly used IIG benchmarks and HUNL Subgames. The experimental results are shown in Figure \ref{fig:PSPCFR+-with-other-algorithms-classical-cfr} and Table \ref{tab:PSPCFR+-with-other-algorithms-HUNL Subgames-classical-cfr}, respectively. We observed that APCFR$^+$ and SAPCFR$^+$ consistently outperformed CFR and CFR$^+$ across all games. However, in poker games like Leduc Poker and HUNL Subgames, APCFR$^+$ and SAPCFR$^+$ did not surpass DCFR. Notably, our algorithms and DCFR are not mutually exclusive and can be combined effectively. Specifically, the core idea of our algorithm—the asymmetry of step sizes—can be integrated with that of DCFR, which discounts prior iterations when calculating accumulated regrets. Consequently, we apply the asymmetry mechanism for step sizes to enhance PDCFR$^+$~\citep{xu2024minimizing}, a combination of DCFR and PCFR$^+$, resulting in APDCFR$^+$. We only make evaluation in HUNL Subgames as Leduc Poker is extremely smaller than HUNL Subgames. The experimental results are presented in Table \ref{tab:PSPCFR+-with-other-algorithms-HUNL Subgames-classical-cfr}. APDCFR$^+$ significantly outperforms both PDCFR$^+$, APCFR$^+$, SAPCFR$^+$, and DCFR (note that APCFR$^+$ and SAPCFR$^+$ outperform PDCFR$^+$). We also compare APDCFR$^+$ with DCFR$^+$~\citep{xu2024minimizing}, an advanced variant of DCFR. The experimental results in Table \ref{tab:PSPCFR+-with-other-algorithms-HUNL Subgames-classical-cfr} indicates that APDCFR$^+$ also significantly outperforms DCFR$^+$. 


As mentioned earlier in our main text, we further evaluate the performance of DCFR, PCFR$^+$, APCFR$^+$, SAPCFR$^+$, DCFR$^+$, and APDCFR$^+$ on Leduc Poker variants, which is larger than the vanilla Leduc Poker. Detailed experimental results and discussions can be found in the final paragraph of Section \ref{sec:experiments} in the main text.

\section{Limitations}
\noindent The primary limitation of APCFR$^+$ and SAPCFR$^+$ is the theoretical convergence rate of $O(1/\sqrt{T})$. Notably, nearly all CFR algorithms exhibit a theoretical convergence rate of $O(1/\sqrt{T})$. Only a few CFR algorithms have a theoretical convergence rate exceeding $O(1/\sqrt{T})$, \eg, Reg-CFR and Clairvoyant CFR, which achieve $O(1/T^{\frac{3}{4}})$ and $O(1/T)$ theoretical convergence rates, respectively. Unfortunately, as shown in our experiments, the empirical convergence rate of these algorithms fails to surpass that of CFR algorithms, which have a theoretical convergence rate of $O(1/\sqrt{T})$, \eg, PCFR$^+$. Interestingly, to achieve a faster empirical convergence rate, some CFR algorithms like DCFR, intentionally compromise their theoretical convergence rate. Specifically, the theoretical convergence rate of DCFR is 3 times slower than that of CFR$^+$ (under the same number of iterations, the theoretical upper bound on the exploitability of DCFR is three times higher than that of CFR$^+$), as shown in Theorem 3 of \citet{tammelin2015solving} (CFR$^+$) and Theorem 2 of \citet{brown2019solving} (DCFR), while DCFR usually outperforms CFR$^+$, as shown in our experiments.


\begin{table}[t]  
\centering %
\begin{tabular}{l@{\hskip 4pt}cccc}  
\toprule  
& PCFR$^+$ & APCFR$^+$ & SAPCFR$^+$ & APCFR$^+$ V2 \\
\midrule  
S3 & \texttt{1.44e-3}  & \texttt{1.02e-3} (\textcolor{red}{-29.2\%}) & \texttt{9.44e-4} (\textcolor{red}{-34.4\%}) & \texttt{9.84e-4} (\textcolor{red}{-31.7\%}) \\
S4 & \texttt{1.04e-3}  & \texttt{7.53e-4} (\textcolor{red}{-27.6\%}) & \texttt{7.83e-4} (\textcolor{red}{-24.7\%}) & \texttt{7.38e-4} (\textcolor{red}{-29.0\%}) \\
\bottomrule  
\end{tabular}  
\caption{The final exploitability for PCFR$^+$, APCFR$^+$, SAPCFR$^+$, and APCFR$^+$ V2 in HUNL Subgames. Values in \textcolor{red}{red} indicate percentages relative to PCFR$^+$.}  
\label{fig:PSPCFR+ V2-with-other-algorithms-HUNL Subgames}  
\end{table}

\begin{table}[t]  
\centering %
\begin{tabular}{l@{\hskip 4pt}cccccc}  
\toprule  
& CFR & CFR$^+$ & \quad DCFR \quad & \quad APCFR$^+$ \quad & \textcolor{white}{x} SAPCFR$^+$ \textcolor{white}{x}  & PDCFR$^+$ \\
\midrule  
S3 & \texttt{1.96e-2} & \texttt{1.16e-2} & \texttt{3.05e-4} & \texttt{1.02e-3} & \texttt{9.44e-4} & \texttt{1.08e-3} \\
S4 & \texttt{2.68e-2} & \texttt{1.82e-2} & \texttt{2.18e-4} & \texttt{7.53e-4} & \texttt{7.83e-4} & \texttt{1.10e-3} \\
\midrule  
& DCFR$^+$ & APDCFR$^+$ & & & \\
\midrule  
S3 & \texttt{2.23e-4} (\textcolor{blue}{-26.8\%}) & \texttt{1.69e-4} (\textcolor{blue}{-44.6\%}) & & & \\
S4 & \texttt{1.55e-4} (\textcolor{blue}{-28.9\%}) & \texttt{1.27e-4} (\textcolor{blue}{-41.7\%}) & & & \\
\bottomrule  
\end{tabular}  
\caption{The final exploitability for CFR, CFR$^+$, and APDCFR$^+$ in HUNL Subgames. Values in \textcolor{blue}{blue} indicate percentages relative to DCFR. ``S3" and ``S4" denote Subgame 3 and Subgame 4, respectively.} 
\label{tab:PSPCFR+-with-other-algorithms-HUNL Subgames-classical-cfr}  
\end{table}

\end{document}